 \documentclass[final,12pt]{clear2025} % Include author names

\newcommand{\thm}{\begin{theorem}}
\newcommand{\lem}{\begin{lemma}}
\newcommand{\pro}{\begin{proposition}}
\newcommand{\dfn}{\begin{definition}}
\newcommand{\rem}{\begin{remark}}
\newcommand{\xam}{\begin{example}}
\newcommand{\cor}{\begin{corollary}}
\newcommand{\prf}{\noindent{\bf Proof:} }
\newcommand{\ethm}{\end{theorem}}
\newcommand{\elem}{\end{lemma}}
\newcommand{\epro}{\end{proposition}}
\newcommand{\edfn}{\bbox\end{definition}}
\newcommand{\erem}{\bbox\end{remark}}
\newcommand{\exam}{\bbox\end{example}}
\newcommand{\ecor}{\end{corollary}}
\newcommand{\eprf}{\bbox\vspace{0.1in}}
\newcommand{\beqn}{\begin{equation}}
\newcommand{\eeqn}{\end{equation}}
\newcommand{\bbox}{\vrule height7pt width4pt depth1pt}

\newcommand{\clm}{\begin{claim}}
\newcommand{\eclm}{\end{claim}}
\newcommand{\sat}{\models}

\newcommand{\rimp}{\Rightarrow}
\newcommand{\union}{\cup}
\newcommand{\xx}{\vec{x}}
\newcommand{\yy}{\vec{y}}

\renewcommand{\phi}{\varphi}
\newcommand{\F}{{\cal F}}
\newcommand{\G}{{\cal G}}

\renewcommand{\P}{{\cal P}}

\newcommand{\R}{{\cal R}}

\newcommand{\U}{{\cal U}}
\newcommand{\V}{{\cal V}}

\newcommand{\<}{\langle}
\renewcommand{\>}{\rangle}

\newcommand{\ol}{\setlength{\itemsep}{0pt}\begin{enumerate}}
\newcommand{\eol}{\end{enumerate}\setlength{\itemsep}{-\parsep}}
\newcommand{\ul}{\setlength{\itemsep}{0pt}\begin{itemize}}
\newcommand{\dl}{\setlength{\itemsep}{0pt}\begin{description}}
\newcommand{\edl}{\end{description}\setlength{\itemsep}{-\parsep}}
\newcommand{\eul}{\end{itemize}\setlength{\itemsep}{-\parsep}}

\newcommand{\true}{{\it true}}

\usepackage{amssymb}    

\renewcommand{\S}{{\cal S}}

\newtheorem*{theorem*}{Theorem}
\newtheorem*{proposition*}{Proposition}
\newtheorem*{example*}{Example}

\newcommand{\YY}{\vec{Y}}
\newcommand{\XX}{\vec{X}}
\newcommand{\ZZ}{\vec{Z}}
\newcommand{\zz}{\vec{z}}

\title[Nondeterministic Causal Models]{Nondeterministic Causal Models}
\usepackage{times}
\clearauthor{%
 \Name{Sander Beckers} \Email{srekcebrednas@gmail.com}\\
 \addr Cornell University
}

\begin{document}

\maketitle

\begin{abstract}%
  I generalize acyclic deterministic structural causal models to the nondeterministic case and argue that this offers an improved semantics for counterfactuals. The standard, deterministic, semantics developed by Halpern (and based on the initial proposal of Galles \& Pearl) assumes that for each assignment of values to parent variables there is a unique assignment to their child variable, and it assumes that the actual world (an assignment of values to all variables of a model) specifies a unique counterfactual world for each intervention. Both assumptions are unrealistic, and therefore I drop both of them in my proposal. I do so by allowing {\em multi-valued} functions in the structural equations. In addition, I adjust the semantics so that the solutions to the equations that obtained in the actual world are preserved in any counterfactual world. I provide a sound and complete axiomatization of the resulting logic and compare it to the standard one by Halpern and to more recent proposals that are closer to mine. Finally, I extend these models to the probabilistic case and show that they open up the way to identifying counterfactuals even in Causal Bayesian Networks.
\end{abstract}


\section{Introduction}\label{sec:int}

Deterministic Structural Causal Models -- DSCMs from now on -- represent the causal relations between a set of endogenous variables by specifying an equation for each endogenous variable that determines the variable's value as a function of the values of some other variables, both endogenous ($\V$) and exogenous ($\U$) \citep{pearl:book2}. Exogenous variables represent unobserved sources of variation whose existence has to be assumed in order to obtain {\em deterministic} equations. 

DSCMs serve as the mathematical and conceptual foundation for Pearl's causal modelling approach. As they lack probabilities, probabilistic causal models are built on top of DSCMs in two stages. First, we get probabilistic DSCMs -- PDSCMs -- by adding a probability distribution over the exogenous variables $\U$. This induces a joint distribution over $(\U,\V)$. Second, assuming that there are no cyclic causal relations, that exogenous variables are independent, and that no two endogenous variables share an exogenous parent (i.e., that the model is {\em Markovian}), we get {\em Causal Bayesian Networks} by marginalizing out the exogenous variables and directly considering the marginal distribution over $\V$ and its Markov factorization \citep[Ch.1]{pearl:book2}\citep[Th.2]{bareinboim22}.

I here present {\em Nondeterministic Structural Causal Models} -- NSCMs -- as a more general and improved foundation for causal models. The reason for doing so is that the heavy reliance on exogenous variables in DSCMs is unnecessarily restrictive. The generalization from DSCMs to NSCMs is given by dropping the assumption that there must exist exogenous variables such that the value of each endogenous variable can be uniquely determined by its causal parents. Note that dropping this assumption is not the same as dropping the assumption that the world is itself fundamentally deterministic. It is one thing to assume that there must exist some properties of the world such that the value of a variable is uniquely determined, it is quite another to assume that these properties can be neatly compartmentalized into sets of variables whose values determine the other value according to a stable functional relationship. In many situations the latter may not hold, and such situations are currently beyond the scope of SEMs. (And in even more situations, we may want to -- or have to due to practical constraints -- represent the world {\em as if} it does not hold.)

Furthermore, the use of exogenous variables also involves a commitment to an overly strong semantics for causal counterfactual statements, for it implies that there is always a {\em unique} solution to the model for any counterfactual query, given the actual values of all variables. That assumption is even stronger than the previous one. For example, consider a patient who may or may not receive treatment for some condition, and when she does she may or may not recover, depending on whether the treatment is effective. We observe that in the actual world, she does not receive treatment, and she does not recover. Representing this world using DSCMs requires specifying {\em actual} values of exogenous variables such that if she had been treated, she would have either certainly recovered, or she would certainly not have. Using NSCMs there is no such requirement, and thus we can simply consider both counterfactual worlds as possible. (See Examples \ref{ex:simple} and \ref{ex:comp} for the formal details.) 

The benefit of dropping this counterfactual {\em uniqueness property} should not be underestimated. The property has been severely criticized in the general philosophical literature on counterfactuals ever since the seminal work of \cite{lewis73a} first did so. More relevant for the present purposes, it is one of the driving forces behind \cite{dawid00}'s influential criticism of the counterfactual semantics offered by both DSCMs and the related {\em Potential Outcomes} framework \citep{rubin74}, and this criticism has resurfaced in the recent back-and-forth between \cite{mueller23} on the one hand and \cite{dawid23} and \cite{sarvet23} on the other over the correct methodology for personalised treatment decisions, such as in the patient example above. 

It is worth quoting Dawid and Senn at length to illustrate how much their animosity towards counterfactuals is based on these two deterministic assumptions: 
\begin{quote}
Indeed, there are serious philosophical objections to regarding potential responses [unique counterfactual outcomes] as having real existence. Only if we take a fully Laplacean view of the universe, in which the future of the universe is entirely determined by its present state and the laws of Physics, does this make any sense at all -- and even then, it is difficult to incorporate the whims of an unconstrained external agent who decides whether or not to give treatment, or to account for the effect of external conditions arising after treatment. Even under Laplacean determinism, our ignorance of the information needed to predict the future means that we are unable to make use of it. Whether or not we believe in a deep-down deterministic universe, our predictions of the future can only be based on the limited information we do have at our disposal, and must necessarily be probabilistic. \citep[p. 5]{dawid23} 
\end{quote}
By offering a semantics for counterfactuals in nondeterministic causal models, I aim to provide room for both sides of this hotly contested -- and literally vital -- dispute to meet in the middle. Although I agree with Dawid and Senn that probabilistic predictions are crucial, I agree with Pearl and \cite{halpernbook} that an understanding of the logic for counterfactuals should precede a probabilistic account of counterfactuals, and hence that is where I start. 

Concretely, I proceed as follows. I define NSCMs (Sec. \ref{sec:cm}), present a formal language and several corresponding semantics for causal formulas that hold in NSCMs (Sec. \ref{sec:lan}), compare them to the standard one for DSCMs (Sec. \ref{sec:com}), offer a sound and complete axiomatization for two of the novel logics and compare them to other recent proposals that allow nondeterminism (Sec. \ref{sec:rel}), and offer a preliminary investigation into the probabilistic generalization of NSCMs, illustrating how it allows for computing counterfactuals even in Causal Bayesian Networks (Sec. \ref{sec:prob}). %We leave a full investigation of probabilistic causal models for future work. 

\section{Nondeterministic Structural Causal Models}\label{sec:cm}

I take the definition of deterministic causal models by \cite{halpernbook} and generalize it to the nondeterministic case by using multi-valued functions.\footnote{\cite{halpern00} already suggested this generalization, but never implemented it. He did recently offer an even further generalization in order to allow for an infinite number of variables with infinite ranges, by doing away with equations altogether \citep{peters21,halpern22}. Recently \cite{barbero23} and \cite{wysocki24} have likewise offered generalizations to the nondeterministic case. I compare my approach to these ones in Section \ref{sec:rel}.
} As a first step, we need to define a signature as the variables out of which a causal model is built up. 
\dfn
A signature $\cal S$ is a tuple $(\U,\V,\R)$, where $\U$
is a set of \emph{exogenous} variables, $\V$ is a set 
of \emph{endogenous} variables,
and $\R$ a function that associates with every variable $Y \in  
\U \union \V$ a nonempty set $\R(Y)$ of possible values for $Y$
(i.e., the set of values over which $Y$ {\em ranges}).
If $\vec{X} = (X_1, \ldots, X_n)$, $\R(\vec{X})$ denotes the
Cartesian product $\R(X_1) \times \cdots \times \R(X_n)$. \edfn
A causal model expresses the causal relations between the endogenous variables of a signature. In addition to using multi-valued functions, I depart from Halpern by explicitly including the causal graph as an element of the causal model. 
\dfn\label{def:cm}
A \emph{causal model} (or a {\em Nondeterministic Structural Causal Model -- NSCM}) $M$ is a triple $(\cal S,\F, \G)$, 
where $\cal S$ is a signature, $\G$ is a directed graph such that there is one node for each variable in $\cal S$, and
$\F$ defines a  function that associates with each endogenous
variable $X$ a \emph{structural equation} $F_X$ giving the possible values of
$X$ in terms of the 
values of some of the other endogenous and exogenous variables. A structural equation $F_X$ takes on the form $X = f_X(\vec{Pa_X})$, where $\vec{Pa_X} \subseteq (\U \cup \V - \{X\})$ are the {\em parents} of $X$ as they appear in $\G$, and $f_X: \R(\vec{Pa_X}) \rightarrow \P(\R(X))$. (Here $\P(\R(X))$ is the {\em powerset} of $\R(X)$: the set that contains as its elements all subsets of $\R(X)$.)\edfn
We here restrict attention to the case in which $\G$ is acyclic (so that $\G$ is a DAG -- a Directed Acyclic Graph). If for each $f_X$ the co-domain does not contain the empty set, we say that a causal model is {\em total}. We here restrict ourselves to total causal models. This amounts to the assumption that for all possible settings $\vec{pa_X}$ of the parents, there exists at least one solution $x$ for the child.

There are no equations for exogenous variables $\U$, as these are taken to represent the background conditions that are simply given. We call $\vec{u} \in \R(\U)$ a {\em context}, $\vec{v} \in \R(\V)$ a {\em state}, and $(\vec{u},\vec{v}) \in \R(\U \cup \V)$ is a {\em world}. In {\em deterministic} causal models all the functions $f_X$ are standard as opposed to multi-valued, and thus each equation has a unique solution $x$ for each choice of values $\vec{pa_X}$. In nondeterministic models, a {\em solution} of the equation $X=f_X(\vec{Pa_X})$ is a tuple $(x,\vec{pa_X})$ such that $x \in f_X(\vec{pa_X})$. (The term ``equation'' is thus somewhat strange, but we stick with it given the common reference to causal models as ``structural equation models''. Furthermore, as pointed out to me by Joris Mooij, an equivalent characterization can be given in terms of a literal equation as well.\footnote{Concretely, this can be done using the approach of \cite{bongers21}. Although they do not explicitly consider nondeterministic models, their models do allow ``self-cycles'', meaning they allow for $X$ to depend on $X$. Note that $X=f_X(\vec{Pa_X})$ is equivalent to $X=X + (1-1_{f_X}(X,\vec{Pa_X}))$, where $1_{f_X}$ is the indicator function that returns $1$ if $X \in f_X(\vec{Pa_X})$ and $0$ otherwise. This correspondence opens up the possibility of integrating both approaches.}
\iffalse
\footnote{Concretely, it turns out that the approach of \cite{bongers21} can be used to express my nondeterministic ``equations'' as literal equations. Although they do not explicitly consider nondeterministic models, their models do allow ``self-cycles'', meaning they allow for $X$ to depend on $X$. Note that $X=f_X(\vec{Pa_X})$ is equivalent to $X=X + (1-1_{f_X}(X,\vec{Pa_X}))$, where $1_{f_X}$ is the indicator function that returns $1$ if $X \in f_X(\vec{Pa_X})$ and $0$ otherwise. This correspondence opens up the possibility of integrating both approaches.%, but that is something I defer to future work.
 }) 
 \fi
 A solution of $M$ is a world $(\vec{u},\vec{v})$ that is a solution of all equations in $\F$.

In acyclic models the solutions of $M$ given a context $\vec{u}$ can be determined recursively: determine the solutions of each equation in the partial order given by $\G$, and pass these on to the next equation. In general, deterministic causal models (DSCMs) need not have solutions for each context, nor do any such solutions have to be unique, but unicity and existence are guaranteed for acyclic deterministic models. When we move to nondeterministic causal models, unicity and existence are no longer guaranteed even for acyclic models, but existence is recovered in total models. 

%Lastly, we point out that, contrary to standard causal models, we do not really need exogenous variables: making an exogenous variable $U \in \U$ endogenous and giving it the equation $U = \R(U)$ results in what is basically the same model. We keep exogenous variables anyway because conceptually the distinction remains useful.

\section{The Causal Language}\label{sec:lan}

Given a signature $\cal S = (\U,\V,\R)$, an \emph{atomic formula} is a
formula of the form $X = x$, for  $X \in \V$ and $x \in \R(X)$.  
A {\em basic formula (over $\cal S$)\/} $\phi$ is a Boolean combination of atomic formulas. An \emph{intervention} has the form $\vec{Y} \gets \vec{y}$, where $Y_1, \ldots, Y_k$ are distinct variables in $\V$, and $y_i \in \R(Y_i)$ for each $1 \leq i \leq k$.
 A {\em basic causal formula} has the form  $[Y_1 \gets y_1, \ldots, Y_k \gets y_k] \phi$, where $\phi$ is a basic formula and $Y_1,\ldots,Y_k$ are distinct variables in $\V$. Such a formula is abbreviated as $[\vec{Y} \gets \vec{y}]\phi$. The special case where $k=0$ is abbreviated as $[]\phi$.  Finally, a {\em causal formula} is a Boolean combination of basic causal formulas. The language $\cal{L}(\S)$ that we consider consists of all causal formulas.%space\footnote{As we point out in Section \ref{sec:rel}, a basic formula $\phi$ is equivalent to the causal formula $[]\phi$, and thus this is without loss of generality, just as was the case for \cite{halpern00}.}
 
A causal formula $\psi$ is true or false in a causal model, given a world. We write $(M,\vec{u},\vec{v}) \sat \psi$  if the causal formula $\psi$ is true in causal model $M$ given world $(\vec{u},\vec{v})$. We call a model-world pair $(M,\vec{u},\vec{v})$ a {\em causal setting}.

We first define the $\sat$ relation for basic formulas. $(M,\vec{u},\vec{v}) \sat X=x$ if 
%$(\vec{u},\vec{v})$ is a solution of $M$ and 
$x$ is the restriction of $\vec{v}$ to $X$. (From now on we abuse notation and write this simply as $x \in \vec{v}$, with the understanding that lower-case letters refer to values of the variables with that name. Similarly, we write $\vec{x} \subseteq \vec{v}$ whenever we want to allow that $\vec{X}$ consists of multiple variables.) We extend $\sat$ to basic formulas $\phi$ in the standard way. Note that the truth of basic formulas is determined solely by the state $\vec{v}$, and thus we often also write $\vec{v} \sat \phi$. 

In order to define the $\sat$ relation for causal formulas, we introduce two operations on a causal model, the {\em actualized refinement} that is the result of integrating the actual behavior of the equations as observed in a world $(\vec{u},\vec{v})$ into the equations of a model $M$, and the {\em intervened} model that is the result of performing an intervention on the equations of a model $M$.
\dfn\label{def:ref} Given causal models $M'$ and $M$ over identical signatures $\S$ and with identical graphs $\G$, we say that {\em $M'$ is a refinement of $M$} if for all $X \in \V$ and all $\vec{pa_X} \in \R(\vec{Pa_X})$: $f^{M'}_X(\vec{pa_X}) \subseteq f^{M}_X(\vec{pa_X})$.
\edfn
Since we are restricting ourselves to total causal models, the only refinement of a deterministic model is itself. (Concretely, if we drop totality, then a refinement allows for $f_X(\vec{pa_X}) =\emptyset$, which does not result in a deterministic model.) Hence we say that a deterministic model is {\em maximally refined}. Concretely, since the way that the equations determine the outcome in a deterministic model is identical across each world $(\vec{u},\vec{v})$, there is no need to additionally consider how the equations behave in a specific world. This is no longer true for nondeterministic models, as there the actual values that obtained in a world inform us about how the equations behaved for those values (in that specific world). When evaluating formulas in a world, we need to take this information into account, and to do so requires refining the equations so that they incorporate this actual behavior.
\dfn\label{def:ar} Given a solution $(\vec{u},\vec{v})$ of a model $M = (\cal S,\F,\G)$, we define the {\em actualized refinement} $M^{(\vec{u},\vec{v})}$ as the refinement of $M$ in which $\F$ is
replaced by $\F^{(\vec{u},\vec{v})}$, as follows: for each variable $X \in \V$, its function $f_X$  is replaced by $f^{(\vec{pa_X},x)}_X$ that behaves identically to $f_X$ for all inputs except for $\vec{pa_X}$. Instead, $f^{(\vec{pa_X},x)}_X(\vec{pa_X})=x$, where $(x,\vec{pa_X}) \subseteq (\vec{u},\vec{v})$.
\edfn
Setting the value of some variables $\vec{Y}$ to $\vec{y}$ in a causal 
model $M = (\cal S,\F,\G)$ results in a new causal model, denoted $M_{\vec{Y}
\gets \vec{y}}$, which is identical to $M$, except that $\F$ is
replaced by $\F^{\vec{Y} \gets \vec{y}}$: for each variable $X \notin
  \vec{Y}$, $F^{\vec{Y} \gets \vec{y}}_X = F_X$ (i.e., the equation
  for $X$ is unchanged), while for
each $Y'$ in $\vec{Y}$, the equation $F_{Y'}$ for $Y'$ is replaced by $Y' = y'$
(where $y' \in \vec{y}$). Similarly, $\G$ is replaced with $\G^{\vec{Y} \gets \vec{y}}$.

With these operations in place, we can define the $\sat$ relation for basic causal formulas, relative to settings  $(M,\vec{u},\vec{v})$ such that $(\vec{u},\vec{v})$ is a solution of $M$. $(M,\vec{u},\vec{v}) \sat [\vec{Y} \gets \vec{y}]\phi$ iff $\vec{v}' \sat \phi$ for all states $\vec{v}'$ such that $(\vec{u},\vec{v}')$ is a solution of $(M^{(\vec{u},\vec{v})})_{\vec{Y} \gets \vec{y}}$. We inductively extend the semantics to causal formulas in the standard way, that is,  $(M,\vec{u},\vec{v}) \sat [\vec{Y} \gets \vec{y}]\phi_1 \land [\vec{Z} \gets \vec{z}]\phi_2$ iff $(M,\vec{u},\vec{v}) \sat [\vec{Y} \gets \vec{y}]\phi_1$ and $(M,\vec{u},\vec{v}) \sat [\vec{Z} \gets \vec{z}]\phi_2$, and similarly for $\lnot $ and $\lor$.

We define $\langle \vec{Y} \gets \vec{y} \rangle \phi$ as an abbreviation of  $\neg [\vec{Y} \gets \vec{y}] \neg \phi$. So $(M,\vec{u},\vec{v}) \sat \langle \vec{Y} \gets \vec{y} \rangle \true$ iff there is some world $(\vec{u},\vec{v}')$ that is a solution of $(M^{(\vec{u},\vec{v})})_{\vec{Y} \gets \vec{y}}$. We then write $(M,\vec{u},\vec{v}) \sat \langle \vec{Y} \gets \vec{y} \rangle \V=\vec{v}'$.

We can also evaluate formulas with respect to just a partial causal setting $(M,\vec{u})$, or even with respect to a model $M$ by itself. For basic causal formulas, we define that $(M,\vec{u}) \sat  [\vec{Y} \gets \vec{y}]\phi$ iff $(M,\vec{u},\vec{v}) \sat  [\vec{Y} \gets \vec{y}]\phi$ holds for all states $\vec{v}$ such that $(\vec{u},\vec{v})$ is a solution of $M$. In a similar fashion, we define that $M \sat  [\vec{Y} \gets \vec{y}]\phi$ iff $(M,\vec{u},\vec{v}) \sat  [\vec{Y} \gets \vec{y}]\phi$ holds for all solutions $(\vec{u},\vec{v})$ of $M$. %, and similarly for $(M,\vec{v}) \sat \psi$ with the roles of $\vec{u}$ and $\vec{v}$ reversed. %Note: think about this more: (The generalization to subtuples $\vec{X} \subseteq \U \cup \V$ and settings $(M,\vec{x}) \sat [\vec{Y} \gets \vec{y}]\phi$  is straightforward.)
We again inductively extend to causal formulas in the standard way: $(M,\vec{u}) \sat [\vec{Y} \gets \vec{y}]\phi_1 \land [\vec{Z} \gets \vec{z}]\phi_2$ iff $(M,\vec{u}) \sat [\vec{Y} \gets \vec{y}]\phi_1$ and $(M,\vec{u}) \sat [\vec{Z} \gets \vec{z}]\phi_2$, and similarly for $\lnot $ and $\lor$. Likewise for $M \sat \psi$. 

We have now defined three different semantics for $\cal{L}(\S)$: the first with respect to full causal settings $(M,\vec{u},\vec{v})$, the second with respect to partial causal settings $(M,\vec{u})$, and the third with respect to $M$. The first evaluates causal formulas relative to a single world $(\vec{u},\vec{v})$, and thus we will call the resulting logic the {\em single world counterfactual logic}, or {\bf swc} logic for short. It formalizes counterfactual statements relative to a specific {\em actual world}. The second only requires specifying a single context $\vec{u}$, and thus it considers the entire set of worlds $(\vec{u},\vec{v})$ that extend $\vec{u}$ together, and thus we call the resulting logic the {\em single context counterfactual logic}, or {\bf scc} logic for short. Lastly, the third semantics evaluates formulas without restricting the possible worlds any further than the model $M$ already does, and thus we call the resulting logic the {\em single model counterfactual logic}, or {\bf smc} logic for short. In the remainder of this paper our attention goes to the first two counterfactual logics, but I flag the {\bf smc} logic as a worthwhile subject for future investigation.

\section{Comparison to the Standard Semantics}\label{sec:com}

To recap, the standard logic for causal models by \cite{halpern00,halpernbook} is a logic for DSCMs instead of NSCMs, and its semantics is defined identically to mine except that it does not use the actualized refinement. Given that for any DSCM the actualized refinement $M^{(\vec{u},\vec{v})}$ will simply be $M$ anyway, the two semantics are in fact entirely equivalent for DSCMs. 

Concretely, the standard semantics defines the $\sat$ relation for basic causal formulas as: $(M,\vec{u},\vec{v}) \sat [\vec{Y} \gets \vec{y}]\phi$ iff $\vec{v}' \sat \phi$ for all states $\vec{v}'$ such that $(\vec{u},\vec{v}')$ is a solution of $M_{\vec{Y} \gets \vec{y}}$. Note that, importantly, this semantics does not depend on the actual state $\vec{v}$, and therefore it can be (and usually is) written and interpreted as a semantics for partial causal settings $(M,\vec{u})$. As a result, the distinction between the counterfactual logics {\bf swc} and  {\bf scc} collapses in the case of DSCMs, and thus there is only a single counterfactual logic.

If the DSCMs are acyclic, then for each $\vec{u}$ there is a unique $\vec{v}$ such that $(\vec{u},\vec{v})$ is a solution of $M$, and thus the counterfactual logic is with respect to a single actual world, as is the case for our {\bf swc}. Also, for each $\vec{u}$, there is a unique $\vec{v}$ such that $(\vec{u},\vec{v})$ is a solution of $M_{\vec{Y} \gets \vec{y}}$ for any $\vec{Y} \gets \vec{y}$, and hence the standard semantics for acyclic DSCMs satisfies the controversial uniqueness property we mentioned in Section \ref{sec:int}. If the DSCMs are cyclic, then for each context there may be multiple solutions, or none. Thus, in this case the counterfactual logic is with respect to the set of worlds that extend a context $\vec{u}$, as is the case for our {\bf scc} (except that for {\bf scc} a solution is guaranteed to exist, given totality). As I discuss in Section \ref{sec:rel}, these semantics still satisfy a property that is conceptually very similar to the uniqueness property. 

Importantly, as the following Theorem shows, when we move from DSCMs to NSCMs, the actualized refinement {\em only matters for {\bf swc}}, and thus for both the {\bf scc} and {\bf smc} logics my semantics are simply generalizations of the standard semantics to the nondeterministic case.
\thm\label{thm:mod} Given a nondeterministic causal model $M$, we have that for all $\vec{Y} \subseteq \V$, for all $\vec{y} \in \R(\vec{Y})$, and for all basic formulas $\phi$:
\begin{itemize}
\item $M \sat [\vec{Y} \gets \vec{y}]\phi$ iff $\vec{v} \sat \phi$ for all solutions $(\vec{u},\vec{v})$ of $M_{\vec{Y} \gets \vec{y}}$.
\item For all contexts $\vec{u}$: $(M,\vec{u}) \sat [\vec{Y} \gets \vec{y}]\phi$ iff $\vec{v} \sat \phi$ for all states $\vec{v}$ such that $(\vec{u},\vec{v})$ is a solution of $M_{\vec{Y} \gets \vec{y}}$.
\end{itemize}
\ethm 
\prf All proofs are to be found in the Appendix. \eprf

The intuition behind Theorem \ref{thm:mod} is that the actualized refinement operation updates $M$'s equations to include their {\em actual behavior}, and there is no actual behavior unless one specifies an actual world $(\vec{u},\vec{v})$, so the operation can be ignored for the {\bf scc} and {\bf smc} logics. %Concretely, single world counterfactuals are statements about what is true in a particular, fully specified, world $(\vec{u},\vec{v})$ that is governed by a causal model $M$, if we were to intervene to set some variables counter to fact.\footnote{Usually one speaks of counterfactuals even when variables are set to their actual values by an intervention, but in such cases the intervention should have no impact on the truth of factual formulas, and Proposition \ref{pro:act} shows that this is indeed the case for my semantics.} Single context counterfactuals, on the other hand, are statements about what holds in all worlds that share a context $\vec{u}$ and are governed by the causal model $M$. Single model counterfactuals are of even wider scope, as they do not rely on any assumption except that the worlds are governed by the model $M$. In both of the latter cases, this means that we take into account {\em all} the possible ways in which $M$ (resp. $(M,\vec{u})$) can be actualized, and thus the various possible actualized refinements ought to cancel each other out. I take the fact that my semantics bears out this intuition to be an important sanity check.

As far as I am aware, the distinction between {\bf swc} and {\bf scc} logics has gone unnoticed so far. In fact, recently \cite{halpern22} introduced generalized structural equation models -- GSEMs -- and they provide an axiomatization for the more general logic that is the result, but in doing so Halpern has moved entirely from an {\bf swc} type of logic -- that was the subject matter of \citep{halpern00} and \citep{galles98} -- towards a {\bf scc} type of logic, for the logic for GSEMs does not allow evaluating formulas with respect to a single world. In the next section I conclude the definition of the {\bf swc} and {\bf scc} logics by offering a sound and complete axiomatization for both, and interpret the results in light of other recently proposed logics for nondeterministic causal reasoning.

\section{Axiomatization and Recent Related Work}\label{sec:rel}

Throughout this section we hold fixed some finite signature $\S=(\U,\F,\R)$, i.e., $\U$ and $\V$ are finite, and $\R(X)$ is finite for all $X \in \U \cup \V$. Let $AX$ be the axiom system for the language $\cal{L}(\S)$ that consists of the following list of axioms and inference rule MP. 
\begin{itemize}
	\item[D0.] All instances of propositional tautologies.
\item[D1.] $[\YY \gets \yy](X = x \rimp
		      X \ne x')$  if $x, x' \in \R(X)$, $x \ne x'$ \hfill
\hfill
(functionality)
	\item[D2.] $[\YY \gets \yy](\bigvee_{x \in \R(X)} X = x)$
\hfill (definiteness)
	\item[D3(a).] $\<\XX \gets \xx\>(W = w
		      \land \phi)
		      \rimp \<\XX \gets \xx, W \gets
		      w\>(\phi)$ if $W \notin \XX$
	      \hfill
(weak composition)
\item[D3(b).] $[\XX \gets \xx](W = w
		      \land \phi)
		      \rimp [\XX \gets \xx, W \gets
		      w](\phi)$ if $W \notin \XX$
	      \hfill
(strong composition)
	\item[D4.] $[\XX \gets \xx](\XX = \xx)$ \hfill
(effectiveness)
	\item[D5.] $(\<\XX \gets \xx, Y \gets y\> (W = w \land
		      \ZZ = \zz)  \land
		      \<\XX \gets \xx, W \gets w\> (Y = y \land
\ZZ = \zz))$\\
	      $\mbox{ }\ \ \ \rimp \<\XX \gets \xx\> (W
		      = w \land Y = y \land \ZZ = \zz)$ if $\vec{Z}
		      = \V - (\vec{X} \cup \{W,Y\})$
\mbox{ }  \hfill (reversibility)
\item[D6.] $(X_0 \rightsquigarrow X_1 \land \ldots \land X_{k-1} \rightsquigarrow X_k) \rimp \lnot (X_k \rightsquigarrow X_0)$
\hfill 	      (recursiveness)
	\item[D7.] $([\XX \gets \xx]\phi \land [\XX \gets
			      \xx](\phi \rimp \psi)) \rimp  [\XX \gets \xx]\psi$
\hfill 	      (distribution)
	\item[D8.] $[\XX \gets \xx]\phi$ if $\phi$ is a propositional
tautology  \hfill (generalization)
	\item[D9.]
	      $\<\YY \gets \yy\>true \wedge (\<\YY \gets \yy\>\phi
		      \Rightarrow [\YY\gets \yy]\phi)$
	      \ if $\YY= \V$ or, for some $X \in \V$, 
	      $\YY = \V - \{X\}$
\hfill           (unique        	      outcomes for $\V$ and
$\V - \{X\}$)
\item[D10(a).]  $\<\YY \gets \yy\>true$
\hfill (at least one outcome)
\item[D10(b).] $\<\YY \gets \yy\>\phi \Rightarrow [\YY\gets \yy]\phi$
\hfill (at most one outcome)
\item[D10(c).]
$\<\>\phi \Rightarrow [] \phi$
   \hfill
(at most one actual outcome)
	\item[MP.] From $\phi$ and $\phi \rimp \psi$, infer $\psi$
\hfill (modus ponens)
\end{itemize}
Here, $Y \rightsquigarrow Z$ means that $Y \neq Z$ and 
%new
%$$ \vee_{\vec{X} \subseteq \V, \vec{x} \in \R(\vec{X}), y \in \R(Y), z \neq z' \in \R(Z)} (\<\vec{X} \gets \vec{x}\>(Z=z)\land \<\vec{X} \gets \vec{x},Y\gets y\>(Z=z')).$$
$ \vee_{\vec{X} \subseteq \V, \vec{x} \in \R(\vec{X}), y \neq y' \in \R(Y), z \in \R(Z)} (\<\vec{X} \gets \vec{x}, Y \gets y\>(Z=z)\land [\vec{X} \gets \vec{x},Y\gets y'](Z \neq z)).$
 
 %Important: it has to be assumed that $Y \neq Z$ for this to be sensible, because you always have that So self-cycles are in fact allowed by recursivity, and thus it's not the same as acyclicity of the graph! For DSCMs this doesn't matter, but for me it does, because NSCMs can be viewed as non-total acyclic but possibly self-cyclic DSCMs.
 
 \cite{halpern22}  define $\rightsquigarrow$ differently. However, my definition is easily seen to be equivalent to theirs in the presence of D10(b). Note that none of the axioms mentions basic formulas $\phi$. For both the standard semantics and for mine this is without loss of generality, because $\phi$ is easily seen to be equivalent to $[]\phi$. (Alternatively, I could have extended the causal language and add the axiom $\phi \Leftrightarrow [] \phi$, which is the choice made by \cite{barbero23}.) Lastly, note that in the presence of D10(b), $AX$ contains redundant axioms: D3(b) is a consequence of D3(a) and D10(b), D9 is a consequence of D2 and D10(a,b),  D10(c) is a consequence of D10(b), and -- as shown by \cite{halpern22} -- D5 is a consequence of D2, D3, D6, D7, D8, and D10(a,b). 
 
\cite{halpern22} show that $AX$ without D3(b) and D10(c) is a sound and complete axiomatization with respect to acyclic DSCMs that have signature $\S$, and thus the same holds for $AX$. Furthermore, none of the D10 axioms holds for cyclic DSCMs. In fact, \cite{halpern22} show that $AX$ without D3(b), D10(a,b,c), and D6, is a sound and complete axiomatization for cyclic DSCMs. 
 
D10(b) implies the {\em uniqueness} property mentioned earlier. To see why, consider some solution $(\vec{u},\vec{v})$ (or context $\vec{u}$) relative to which we are evaluating formulas, and take $\phi$ to be $\V=\vec{v'}$ for some $\vec{v'} \in \V$. D10(b) then commits us to the claim that if the world $(\vec{u},\vec{v'})$ is possible under the counterfactual supposition that $\vec{Y}$ were $\vec{y}$, then it is the only world that is possible under that supposition. Although cyclic DSCMs do not validate D10(b), they do validate D9, and that is conceptually not at all different: it simply restricts the above claim to the special case in which the counterfactual supposition contains all but one endogenous variable. As Example \ref{ex:simple} illustrates, neither axioms are sound for either the {\bf swc} logic or the {\bf scc} logic, and thus they free causal models from their commitment to this controversial property. (For sake of completeness, note that neither axiom is sound for the {\bf smc} logic either, nor is it for the {\bf smc} logic over DSCMs only.) %The following model offers a simple counterexample.
\begin{example}\label{ex:simple} Consider again our patient, and assume for simplicity that their condition is certain to kill them if untreated but they may recover if treated, and we observe that they are not treated. We can model this with just two binary endogenous variables $X,Y$, no exogenous variables, and a graph  $X \rightarrow Y$. The equation for $Y$ is $Y \in \{0,1\}$ if $X=1$ and $Y=0$ if $X=0$. The equation for $X$ is irrelevant.
Note that D10(b) and D9 -- combined with D10(a) -- demand that one counterfactual outcome is certain, i.e., that $(M,X=0,Y=0) \sat [X \gets 1]Y=1 \lor [X \gets 1]Y=0$. This consequence of uniqueness is called {\em the principle of conditional excluded middle} and its unacceptability is the main source of \cite{lewis73a}'s classical criticism of the uniqueness property. Yet here we have that $(M,X=0,Y=0) \sat \<X \gets 1\>Y=1 \land \<X \gets 1\>Y=0$, expressing that if the patient were treated they might have recovered but also if they were treated they might not have recovered. As for the {\bf scc}, note that similarly $M \sat \<X \gets 1\>Y=1$ and yet $M \not \sat [X \gets 1]Y=1$.
\end{example}
We let $AX^{scc}_{non}$ denote the axiom system D0-D8 and D10(a), and $AX^{swc}_{non}$ denotes $AX^{scc}_{non}$ plus D10(c). The following result shows that the {\bf swc} logic is stronger than the {\bf scc} logic, that both of them are weaker than the counterfactual logic for acyclic DSCMS, and that both of them are incomparable to the counterfactual logic for cyclic DSCMs. 
\thm\label{thm:axiom} $AX^{swc}_{non}$ (resp. $AX^{scc}_{non}$) is a sound and complete axiomatization for the language $\cal{L}(\S)$ with respect to the {\bf swc} logic (resp. the {\bf scc} logic) over acyclic NSCMs that have signature $\S$.
\ethm 
D10(c) is what distinguishes the two logics, as it does not hold for the {\bf scc} logic. Simply consider the model $M$ consisting of a single binary endogenous variable $X$ with an equation such that $X=\{0,1\}$. Then we have that both $M \sat \<\> X=1$ and $M \sat \<\> X=0$ and thus also $M \not \sat [] X=1$. That D10(c) is not sound for the {\bf scc} logic is what prevents the first problematic assumption of acyclic DSCMs mentioned in Section \ref{sec:int}, namely that the values $\vec{u}$ for all exogenous variables {\em uniquely} determine an actual world $(\vec{u},\vec{v})$. (As with D10(b), to see this take $\phi$ to be of the form $\V=\vec{v'}$.) In the context of the {\bf swc} logic, D10(c) simply expresses the sensible property that if the world were as it actually is, then the actual world is the only world possible. To see this, note that for any $(M,\vec{u},\vec{v})$, the only way for $\langle \rangle \V=\vec{v'}$ to hold is when $\vec{v'}=\vec{v}$. 

%\subsubsection{\bf Halpern \& Peters.} 

\subsection{Related Work}

Theorem \ref{thm:axiom} shows that both of the {\bf swc} and {\bf scc} logics are stronger than the one introduced in \citep{halpern22} for so-called GSEMs, {\em Generalized Structural Equation Models}, even when restricting to the case that is most similar to ours (the case in which all interventions are considered, the signature is finite, and the graph is acyclic). Briefly, GSEMs do away with structural equations altogether, and simply define there to be some set of states $\vec{v}$ corresponding to each context-intervention pair ($\vec{u},\vec{Y} \gets \vec{y})$. As a result,  this means that the connection between formulas that hold under some intervention $\vec{X} \gets \vec{x}$ and those that should hold under an extension $\XX \gets \xx, W \gets w$ is lost, and thus the axioms D3 (both of them) and D5 are no longer sound, in addition to losing D9 and D10(a,b,c). (See their Theorem 5.3.) 

%\subsubsection{\bf Barbero.} 

The closest logic to mine is that of \cite{barbero23}, as it also generalizes structural causal models by allowing for multi-valued functions in the equations. (His work formed part of the motivation for the current paper.) Except for the fact that he also considers the cyclic case, there are two main differences with my proposal. 

Firstly, his logic builds on {\em team semantics}, meaning that instead of evaluating formulas with respect to either a single solution $(\vec{u},\vec{v})$ or a single context $\vec{u}$ of a model $M$, he does so with respect to a {\em team}, where a team is any subset of solutions of a model $M$. As a result, the logic is a mixture of all three of my logics: consider a singleton team and causal formulas are effectively single world counterfactual formulas, consider a team that consists of all solutions extending some context $\vec{u}$, and causal formulas are effectively single context counterfactual formulas, consider a team that consists of all solutions, and causal formulas are single model counterfactual formulas, consider a team that belongs to none of these categories and the causal formula is some form of hybrid. 

Second, although his semantics also aims to incorporate the actual behavior of the equations into all counterfactual worlds, his semantics is less demanding than my
 actualized refinements are. It merely requires that the values of all non-descendants of the variables in some intervention $\vec{X} \gets \vec{x}$ remain identical to the actual ones in counterfactual worlds.\footnote{At least, that is the intended semantics. When moving from the cyclic to the acyclic case, Barbero drops the requirement on non-descendants. Personal communication with the author confirms that this was unintended.} It is easy to show that this property also holds for my semantics, i.e., for any variable $Y$ that is not a descendant of any variable in $\vec{X}$, we have for any solution $(\vec{u},\vec{v})$ of $M$ that if $(M,\vec{u},\vec{v}) \sat Y=y$ then $(M,\vec{u},\vec{v}) \sat [\vec{X} \gets \vec{x}]Y=y$.

As a result of the second difference, weak composition -- D3(a) -- comes apart from strong composition -- D3(b) -- for Barbero. Just as with the {\bf swc} and {\bf scc} logics, D3(a) is sound and D10(b) is not. However, D3(b) is no longer sound for his logic. I believe this is an undesirable property. Consider again the model from Example \ref{ex:simple}, and  consider the world $(M,X=1,Y=1)$. Then according to both Barbero and the {\bf swc} and {\bf scc} logics it holds that $(M,X=1,Y=1) \sat []Y=1$, yet for Barbero's logic we do not have that $(M,X=1,Y=1) \sat [X \gets 1]Y=1$, the reason being that for his logic $(M,X=1,Y=1) \sat \<X \gets 1\>Y=0$. As Barbero points out, this results in his logic not validating  {\em conjunction conditionalization}. Yet in the causal inference literature this property is considered fundamental and goes by the name {\em consistency} \citep{robins1987,pearl:book2}. (See \citep{walters13} for a thorough defense of this property in the philosophical literature.) Therefore I view it as an argument in favor of my approach that it {\em does} satisfy it: 
\footnote{For what it's worth, I also believe that Barbero's defense is confused. He discusses exactly this example ($X=1$ is Alice flipping a coin and $Y=1$ is its landing heads) and he is explicit that the formula is a counterfactual statement, yet then he says that it expresses what would happen if Alice flips the coin {\em again}, which is of course not a counterfactual statement about Alice's present coin flip but a prediction about Alice's future coin flip.} 
\pro\label{pro:act} %For a solution $(\vec{u},\vec{v})$ of $M$, 
If $(M,\vec{u},\vec{v}) \sat \vec{X} =\vec{x}  \land \phi $ then $(M,\vec{u},\vec{v}) \sat [\vec{X} \gets \vec{x}] \phi$.
\epro 
As to the first difference, it is not clear to me what the notion of a team is supposed to capture. The most straightforward interpretation would be that it expresses the set of ``possible worlds'' in some sense or other. However, that interpretation conflicts with how teams behave. Again considering Example \ref{ex:simple}, we add that the team consists exclusively of the solution $(X=1,Y=1)$. If we then consider the intervention $X \gets 1$, the team suddenly changes to also including the solution $(X=1,Y=0)$. I fail to understand why interventions that do not change anything should be able to result in changing what worlds we consider possible. Another consequence of his use of teams is that his logic invalidates D10(c), contrary to the {\bf swc} logic.

%\subsubsection{\bf Wysocki.} 
\cite{wysocki24} also recently proposed to generalize structural causal models using multi-valued functions. His logic, however, only considers formulas with respect to an entire causal model $M$, and is thus what I have called a single model counterfactual logic. Therefore his work complements the current work, as we here focus on the other two logics instead. There is one important difference between his logic and my {\bf smc} logic though, namely in how we view the idea of a causal parent.

He follows Halpern's standard approach in that he {\em defines} the parent relation between variables rather than reading it of an additional graph $\G$ that I (and Barbero) have added to the definition of a causal model. Concretely, for DSCMs \cite{halpernbook} takes the equation for a variable $Y$ to consist of a function $f_Y: \R(\U \cup \V - \{Y\}) \rightarrow \R(Y)$, and then defines $Y$ {\em depends} on $X$ to mean that there exist $x \neq x' \in \R(X)$ and $\vec{z} \in \R(\U \cup \V - \{X,Y\})$ such that $f_Y(\vec{z},x) \neq f_Y(\vec{z},x')$. Wysocki uses the same definition, except that he generalizes it to the multi-valued case. Both Halpern and Wysocki then take ``$X$ is a parent of $Y$'' to be synonymous with ``$Y$ depends on $X$''. Clearly, if $Y$ depends on $X$ then it will also be a parent in my framework, but the reverse need not hold. In fact, following Barbero, we can distinguish between the graph $\G_M$ that comes with a model and the graph $\G_D$ that is built up out of all the dependence relations, noting that $\G_D$ is a subgraph of $\G_M$.\footnote{As I discuss in the Appendix, axiom D6 corresponds to the property that $\G_D$ is acyclic, which is a consequence of the acyclicity of $\G_M$. It is unclear whether it is possible to also express the acyclicity of $\G_M$ in the causal language.} Although nothing prevents one from restricting my framework to models $M$ in which $\G_M = \G_D$, I believe it is a mistake to enforce this stricter definition of a parent in general, because it prohibits a natural extension to probabilistic causal models. Viewed probabilistically, if conditional on all other variables, $X$ can change the {\em distribution} of $Y$, then $X$ is a parent of $Y$. Changing the distribution does not require than any value $y$ changes from being possible ($p > 0$) to impossible ($p=0$) or vice versa, and yet that is what is required for $Y$ to depend on $X$ (in the non-probabilistic sense defined above). Furthermore, one might want to allow for $X$ being a parent of $Y$ even if it {\em does not} meet this probabilistic requirement, because of a failure of {\em faithfulness} \citep{spirtes00}. This is consistent with my more permissive notion of a parent.

\section{Probabilistic Causal Models}\label{sec:prob}

As mentioned in the introduction, DSCMs give rise to probabilistic DSCMs and -- if Markovian -- these in turn give rise to Causal Bayesian Networks -- CBNs. I here offer an initial study of the nondeterministic counterpart to this construction, starting with a generalization of probabilistic causal models. (Interestingly, \cite{dawid22} already briefly consider this generalization, calling them {\em Stochastic Causal Models}. Unfortunately they quickly dismiss them (see Sec. 12.4) by conflating the problem of offering a semantics for counterfactuals with issues involving actual causation. I discuss the latter within the context of nondeterministic models in \citep{beckers24b}.)

%NOTE: I should also include the condition that the model is causally sufficient in a very strong sense: there are any missing parents of the endogenous variables, not just common parents. The same already holds for the non-probabilistic version though, so I won't discuss it here.

\dfn\label{def:PNSCM}
A \emph{probabilistic causal model} (or {\em a PNSCM}) $M$ is a 4-tuple $(\cal S,\F, \G,P_{\U})$, 
where $\cal S$ is a signature, $\G$ is an acyclic directed graph such that there is one node for each variable in $\cal S$, and
$\F$ defines a function that associates with each endogenous
variable $X$ a family of conditional probability distributions $P_X(X | \vec{Pa_X})$ over $\R(X)$, giving the probability of the values of
$X$ in terms of the {\em parents} $\vec{Pa_X}$ of $X$ as they appear in $\G$. $P_{\U}$ is a probability distribution over $\R(\U)$.\edfn
%A {\em solution} of $P_X$ is a tuple $(x,\vec{pa_X})$ such that $P_X(x | \vec{pa_X}) > 0$. 

A PNSCM is {\em Markovian} if all exogenous variables are mutually independent and do not share a child. We assume that Markovian PNSCMs satisfy the {\em Causal Markov Condition}, which means that the joint distribution $P_M$ over $\R(\U \times \V)$ is given as: $P_M(\U, \V) = \prod_{X \in \V} P_X(X | \vec{Pa_X}) \prod_{U \in \U} P_{\U}(U).$ A {\em solution} of $M$ is a world $(\vec{u},\vec{v})$ such that $P_M(\vec{u},\vec{v}) > 0$. %Throughout the following we restrict attention to Markovian PNSCMs.

%A PNSCM is {\em Markovian} if all exogenous variables are mutually independent and do not share a child. 
The Markov condition continues to hold when we marginalize out the exogenous variables to consider $P_M(\V)$, for we have that $P_M(\V) = \prod_{X \in \V} P^{\V}_X(X | \vec{EPa_X})$, where $\vec{EPa_X}$ are $X$'s endogenous parents and  $P^{\V}_X$ is the result of marginalizing $P_X$ over $X$'s exogenous parents $\vec{XPa_X}$: $P^{\V}_X(X | \vec{EPa_X})=\sum_{\{\vec{xpa_X} \in \R(\vec{XPa_X})\}} P_X(X | \vec{Epa_X}, \vec{xpa_X}) P_{\U}(\vec{xpa_X}).$ (See the Appendix for details.) Therefore we can proceed entirely analogously to the deterministic case and construct Causal Bayesian Networks by taking a Markovian probabilistic causal model and marginalizing over the exogenous variables $\U$ \citep[Th. 1.4.1]{pearl:book2}\citep[Th.2]{bareinboim22}. 

\dfn Given a Markovian PNSCM $M=(\S,\F,\G)$, the {\em Causal Bayesian Network} $C_M$ induced by $M$ is the tuple $(\S_{\V},P_M({\V}),\G_{\V})$, where $\S_{\V}$ and $\G_{\V}$ are obtained from $\S$ and $\G$ by removing the exogenous variables and all of their edges.
\edfn
Per construction, a CBN and its underlying PNSCM have identical {\em observational distributions} $P_M(\V)$, which represent the first layer of Pearl's Causal Hierarchy \citep{bareinboim22}. We arrive at the second layer, that of {\em interventional distributions} $P_{M_{\vec{X} \gets \vec{x}}}(\V)$ (usually written as $P(\V | do(\vec{x})$) in exactly the same manner as in the deterministic case, namely by considering the distribution corresponding to the intervened model $M_{\vec{X} \gets \vec{x}}$:
\begin{equation*}
 P_{M_{\vec{X} \gets \vec{x}}}(\V=\vec{v})= 
  \begin{cases} \prod_{\{Y \in \V \setminus \vec{X}\}} P^{\V}_Y(Y | \vec{EPa_Y}) \text { if } \vec{x} \subseteq \vec{v}\\
  0 \text{ otherwise } \end{cases}
\end{equation*}
Just as in the deterministic case, this distribution only depends on knowledge of $P_M(\V)$ and the graph $\G_{\V}$, and therefore it can be viewed interchangeably as an interventional distribution of $M$ or as an interventional distribution of $C_M$. So far nothing new. However,  the nondeterministic and deterministic case start diverging once we move up to the third and most expressive {\em counterfactual} layer of Pearl's Causal Hierarchy. To get there, we first need to define the probabilistic generalization of the actualized refinement (Def. \ref{def:ar}).

\dfn\label{def:ar2} Given a solution $(\vec{u},\vec{v})$ of a probabilistic model $M = (\cal S,\F,\G,P_{\U})$, we define the {\em actualized refinement} $M^{(\vec{u},\vec{v})}$ as the model  in which $\F$ is
replaced by $\F^{(\vec{u},\vec{v})}$, as follows: for each variable $X \in \V$ and $(x,\vec{pa_X}) \subseteq (\vec{u},\vec{v})$, the family of distributions $P_X$  is replaced with the distributions $P^{(\vec{pa_X},x)}_X(X | \vec{pa_X}')$ that are identical to $P_X(X | \vec{pa_X'})$ for all $\vec{pa_X}' \in \R(\vec{Pa_X})$ except for $\vec{pa_X}$. Instead, $P^{(\vec{pa_X},x)}_X(x' | \vec{pa_X})=1$ if $x'=x$ and $0$ otherwise.
\edfn

Just as for the non-probabilistic case, counterfactual statements are suppositions relative to an intervention $\vec{X} \gets \vec{x}$ that is performed in some actual world $(\vec{u},\vec{v})$, and as before, we formalize them by combining the actual refinement and the intervention operators. The difference compared to the logical setting is that we move from statements regarding either {\em all} solutions of the model $(M^{(\vec{u}, \vec{v})})_{\vec{X} \gets \vec{x}}$ (using $[\vec{X} \gets \vec{x}]$) or {\em some} solutions (using $\<\vec{X} \gets \vec{x}\>$) to probabilistic statements about $(M^{(\vec{u}, \vec{v})})_{\vec{X} \gets \vec{x}}$, the joint distribution of which takes on the form:
\begin{equation*}
 P_{(M^{(\vec{u}, \vec{v})})_{\vec{X} \gets \vec{x}}}(\U=\vec{u}^*,\V=\vec{v}^*)=
  \begin{cases} \prod_{\{Y \in \V \setminus \vec{X}\}} P^{(\vec{pa_Y},y)}_Y(y^* | \vec{pa_Y}^*)P_{\U}(\vec{u}^*) \text { if } \vec{x}^* = \vec{x}\\
  0 \text{ otherwise. } \end{cases}
\end{equation*}
Here $(\vec{pa_Y}^*,y^*,\vec{x}^*) \subseteq (\vec{u}^*,\vec{v}^*)$ and $(\vec{pa_Y},y,\vec{x}) \subseteq (\vec{u},\vec{v})$. If we only consider a specific actual context $\U=\vec{u}$ and apply Definition \ref{def:ar2}, this results in:
\begin{equation*}
 P_{(M^{(\vec{u}, \vec{v})})_{\vec{X} \gets \vec{x}}}(\V=\vec{v}^* | \U=\vec{u})=
  \begin{cases}
  0 \text { if }  \vec{x}^* \neq \vec{x}\\
    0 \text { if }  \emptyset \neq \{Y \in \V \setminus \vec{X} | \vec{epa_Y} = \vec{epa_Y}^* \text{ and } y^* \neq y\}\\
  1\text{ if } \emptyset=\{Y \in \V \setminus \vec{X} | \vec{epa_Y} \neq \vec{epa_Y}^*\}\\
  \prod_{\{Y \in \V \setminus \vec{X} | \vec{epa_Y} \neq \vec{epa_Y}^*\}} P_Y(y^* | \vec{epa_Y}^*,\vec{xpa_Y})  \text{ otherwise. }\end{cases}
\end{equation*}
We write $P_M(\vec{y}_{\vec{x}} | \vec{u},\vec{v})$ as shorthand for $P_{(M^{(\vec{u}, \vec{v})})_{\vec{X} \gets \vec{x}}}(\vec{Y}=\vec{y} | \U=\vec{u})$ for $\vec{Y},\vec{X} \subseteq \V$, which is the probabilistic counterpart of evaluating a basic causal formula in our {\bf swc} logic (Sec. \ref{sec:lan}): $(M,\vec{u},\vec{v}) \sat [\vec{X} \gets \vec{x}]\vec{Y}=\vec{y}$. 

%%%%%%%

We now have all the tools in hand to construct the third layer, consisting of {\em probabilities of counterfactuals}, which are expressions of the form $P_M(\vec{y}_{\vec{x}}, \ldots, \vec{z}_{\vec{w}})$. What makes them counterfactual as opposed to merely interventional is that the various interventions $\vec{x}$, $\vec{w}$ may conflict with each other, in which case their respective outcomes $\vec{y}$ and $\vec{z}$ occur across worlds that are counterfactual to each other. (Note that these expressions  include the basic case in which we intervene after having made an {\em actual} observation, as those can be expressed as $P_M(\vec{y}_{\vec{x}} | \vec{z})=P_M(\vec{y}_{\vec{x}}, \vec{z}) / P_M(\vec{z})$.) Our definition proceeds analogously to the standard deterministic case \citep{bareinboim22}. 

\dfn\label{def:pc} A probabilistic causal model $M$ induces a family of joint distributions over counterfactual events of the form $\vec{y}_{\vec{x}}, \ldots, \vec{z}_{\vec{w}}$ for any $\vec{Y},\vec{X}, \ldots, \vec{Z},\vec{W} \subseteq \V$, as follows:
\begin{align*}
P_M(\vec{y}_{\vec{x}}, \ldots, \vec{z}_{\vec{w}})=
 \sum_{\{\vec{u},\vec{v}\}}  P_M(\vec{y}_{\vec{x}}, \ldots, \vec{z}_{\vec{w}} | \vec{u},\vec{v})P_M(\vec{u}, \vec{v})=
 \sum_{\{\vec{u},\vec{v}\}} P_M(\vec{y}_{\vec{x}} | \vec{u},\vec{v}) \ldots P_M(\vec{z}_{\vec{w}} | \vec{u}, \vec{v}) P_M(\vec{u}, \vec{v}).%$\edfn
\end{align*}
(In the Appendix I show that this probability distribution is indeed well-defined.)\edfn
Intuitively, this definition generalizes the idea that conditional on some actual world $(\vec{u},\vec{v})$, different counterfactual queries $\vec{y}_{\vec{x}}$ and $\vec{z}_{\vec{w}}$ are independent of each other. 

Coming back to our patient example, we can now easily compute the probability of necessity and sufficiency, which \cite{mueller23} here interpret as the {\em probability of benefit} (the probability of harm is similar).\footnote{To be clear, I do not endorse their analysis of harm. Instead I have developed my own analysis of harm and benefit in \citep{beckers23b,beckers24a}. What matters here is that both of our accounts invoke probabilities of counterfactuals.}As we will see, in our -- admittedly very simplistic -- example this probability is point-identifiable from just observational data. Since the lack of identifiability is one of \cite{dawid23}'s main concerns with the former's approach, this illustrates how nondeterministic models can provide room for compromise in the debate over personalized medicine.
\begin{example}\label{ex:comp} Imagine again our patient from before, except that we now also consider it possible that the patient recovers even when untreated. So we get: $P_Y(y | x)=p$ and $P_Y(y | x')=q$ and $P_X(x)=r$ for some $0<p,q,r<1$, where $Y=y$/$Y=y'$ represents recovery/death and $X=x$/$X=x'$ represents treatment/no treatment. Computing the PNS gives:
$P_M(y_x,y'_{x'})=P_Y(y | x)P_Y(y' | x')=p(1-q)$. (See the Appendix for the intermediate steps.)
\end{example}

The following result confirms that the probabilistic setting generalizes the logical one.

\thm\label{thm:gen} Say we have a PNSEM $M$ and a NSEM $M^*$ over the same signature and graph which are {\em consistent} with each other, meaning that for all $X \in \V$ and all $x \in \R(X)$, $\vec{pa_X} \in \R(\vec{Pa_X})$: $P_X(x | \vec{pa_X}) > 0$ iff $x \in f_X(\vec{pa_X})$. Then for
 all worlds $(\vec{u},\vec{v})$ and all $\vec{y}_{\vec{x}}, \ldots, \vec{z}_{\vec{w}}$ for some $\vec{Y},\vec{X}, \ldots, \vec{Z},\vec{W} \subseteq \V$:
\begin{align*}
P_M(\vec{y}_{\vec{x}},\ldots, \vec{z}_{\vec{w}} | \vec{u},\vec{v})=1
%$P_M(\vec{y}_{\vec{x}} | \vec{u},\vec{v}) \ldots P_M(\vec{z}_{\vec{w}} | \vec{u}, \vec{v})=1$
\text{ iff }(M^*,\vec{u},\vec{v}) \sat [\vec{X} \gets \vec{x}]\vec{Y} =\vec{y} \land \ldots \land [\vec{W} \gets \vec{w}]\vec{Z}=\vec{z}.
\end{align*}
\ethm
Furthermore, the following results confirm that Definition \ref{def:pc} also generalizes the standard definition of probabilities of counterfactuals \citep[Defns. 5 and 7]{bareinboim22}. (Note that a standard, deterministic, probabilistic causal model, i.e., a PDSCM, is simply a PNSCM for which all $P_X(X | \vec{Pa_X})$ only take value in $\{0,1\}$, meaning that $P_X$ reduces to a function $f_X$.)

\thm\label{thm:pdscm} Given a PDSCM $M$ and $\vec{Y},\vec{X}, \ldots, \vec{Z},\vec{W} \subseteq \V$, it holds that
\begin{align*}
P_M(\vec{y}_{\vec{x}}, \ldots, \vec{z}_{\vec{w}})= \sum_{\{\vec{u} | (M,\vec{u}) \sat [\vec{X} \gets \vec{x}]\vec{Y} =\vec{y} \land \ldots \land [\vec{W} \gets \vec{w}]\vec{Z}=\vec{z}\}}
P_M(\vec{u}).
\end{align*}
 \ethm

\cor Given a PDSCM $M$ and $\vec{Y},\vec{X} \subseteq \V$, it holds that
%\begin{align*}
$P_M(\vec{y}_{\vec{x}})=P_{M_{\vec{X} \gets \vec{x}}}(\vec{y}).$
%\end{align*}
\ecor

\subsection{Counterfactuals in Causal Bayesian Networks}

Just as in the deterministic case, probabilities of counterfactuals cannot be evaluated directly within CBNs because they depend on knowledge of the full graph $\G$ and the full distribution $P_M(\U, \V)$. However, the nondeterministic case {\em does} open up an entirely novel perspective on procedures for partially identifying such probabilities from partial knowledge about a causal model -- such as knowledge of the CBN it induces, for example -- since all of the standard procedures rely on the assumption that the ground truth causal model has to be a PDSCM \citep{balke94,tian00,zhang22}. I defer a full discussion of this change in perspective to future work, and here restrict attention to just one important observation.

In the deterministic case, a conditional probability $P_X(X | \vec{EPa_X})$ expressing the relation between a variable and its endogenous parents in a CBN is assumed to reduce entirely to an unknown probability $P(\vec{XPa_X})$ over some unknown exogenous variables $\vec{XPa_X}$ and an unknown function $f_X(\vec{EPa_X},\vec{XPa_X})$. So unless all the values of $P_X(X | \vec{EPa_X})$ happen to belong to $\{0,1\}$, this reductive viewpoint forces us to conclude that there {\em must} be some missing variables, i.e., $\vec{XPa_X} \neq \emptyset$. In the nondeterministic case we do not have to make this reductive assumption (although of course nothing prevents us from making it  if there is good reason to). Thus one option that is always available for any CBN is to take the probabilities $P_X$ at face value and simply assume that the CBN already contains all relevant variables, so that the ground truth PNSCM is simply identical to the CBN, i.e, to assume that $\U=\emptyset$. Under this option probabilities of counterfactuals (Def \ref{def:pc}) can be computed directly from the CBN. Of course the assumption that $\U=\emptyset$ is a very strong one and thus it requires careful justification. In future work I intend to investigate much weaker assumptions that nonetheless offer informative results on partial identification. 

Interestingly, since the writing of this article, \cite{galhotra24} have likewise offered a novel semantics for probabilities of counterfactuals that can be applied directly to CBNs, and therefore a comparison with their approach is in order as well. For now I tentatively conjecture that the semantics we develop are equivalent, possibly up to some minor technical details in formulation. 

%%%%
\section{Conclusion}

I here developed a nondeterministic generalization of causal models and offered an axiomatization for two of the resulting counterfactual logics, arguing that their semantics are superior to other proposals. Crucially, they inherit all the benefits of Pearl's framework whilst dropping the controversial uniqueness property.
%There has been a growing literature on the semantics of counterfactuals both within and beyond the causal modeling tradition, and we consider it worth investigating further how our semantics compares to other prominent proposals. 
As many applications require probabilistic causal models, 
I also initiated the probabilistic extension of this framework, but I consider it far from concluded. Among other things, I anticipate that it offers a fruitful middle ground between the two camps disputing the role of counterfactuals in personalised decision-making.
%new
Lastly, in a companion paper \citep{beckers24b} I take advantage of the additional expressivity of nondeterministic causal models to offer a novel and improved account of {\em actual causation}. In doing so I also develop a generalization of the notion of {\em counterfactual dependence} and reflect on the role of both notions for the logic of causal discovery.

% Acknowledgments---Will not appear in anonymized version
\acks{I thank Fausto Barbero, Joe Halpern, Joris Mooij, and Dzelila Siljak, for helpful discussions on some aspects of this work, and I thank the reviewers for their insightful reviews. I am particularly grateful for the pushback I received from the reviewers on the probabilistic setting, because this resulted in a thorough revision of that section. This work was funded by ARO grant: W911NF-22-1-0061.}

\bibliographystyle{ACM-Reference-Format}
\bibliography{allpapers}

\begin{thebibliography}{28}
\providecommand{\natexlab}[1]{#1}
\providecommand{\url}[1]{\texttt{#1}}
\expandafter\ifx\csname urlstyle\endcsname\relax
  \providecommand{\doi}[1]{doi: #1}\else
  \providecommand{\doi}{doi: \begingroup \urlstyle{rm}\Url}\fi

\bibitem[Balke and Pearl(1994)]{balke94}
Alexander Balke and Judea Pearl.
\newblock Counterfactual probabilities: Computational methods, bounds and
  applications.
\newblock In \emph{Proceedings of the 10th Conference on Uncertainty in
  Artificial Intelligence (UAI 1994)}, volume~10, pages 46--54, 1994.

\bibitem[Barbero(2024)]{barbero23}
Fausto Barbero.
\newblock On the logic of interventionist counterfactuals under indeterministic
  causal laws.
\newblock In \emph{International Symposium on Foundations of Information and
  Knowledge Systems (FoIKS 2024)}, pages 203--221, 2024.

\bibitem[Bareinboim et~al.(2022)Bareinboim, Correa, Ibeling, and
  Icard]{bareinboim22}
Elias Bareinboim, Juan~D. Correa, Duligur Ibeling, and Thomas Icard.
\newblock On pearl's hierarchy and the foundations of causal inference.
\newblock In \emph{Probabilistic and Causal Inference: The Works of Judea
  Pearl}, pages 507--556. Association for Computing Machinery, 2022.

\bibitem[Beckers(2025)]{beckers24b}
Sander Beckers.
\newblock Actual causation and nondeterministic causal models.
\newblock In \emph{Proceedings of the 4th Conference on Causal Learning and
  Reasoning (CLeaR 2025)}. PMLR, 2025.

\bibitem[Beckers et~al.(2023{\natexlab{a}})Beckers, Chockler, and
  Halpern]{beckers23b}
Sander Beckers, Hana Chockler, and Joseph~Y. Halpern.
\newblock Quantifying harm.
\newblock In \emph{Proceedings of the Thirty-Second International Joint
  Conference on Artificial Intelligence, {IJCAI-23}}, pages 363--371,
  2023{\natexlab{a}}.

\bibitem[Beckers et~al.(2023{\natexlab{b}})Beckers, Halpern, and
  Hitchcock]{beckers23a}
Sander Beckers, Joseph~Y. Halpern, and Christopher Hitchcock.
\newblock Causal models with constraints.
\newblock In \emph{Proceedings of the 2nd Conference on Causal Learning and
  Reasoning (CLeaR 2023)}, 2023{\natexlab{b}}.

\bibitem[Beckers et~al.(2024)Beckers, Chockler, and Halpern]{beckers24a}
Sander Beckers, Hana Chockler, and Joseph~Y. Halpern.
\newblock A causal analysis of harm.
\newblock \emph{Minds and Machines}, 34\penalty0 (3):\penalty0 34, 2024.

\bibitem[Bongers et~al.(2021)Bongers, Forr{\'e}, Peters, and Mooij]{bongers21}
Stephan Bongers, Patrick Forr{\'e}, Jonas Peters, and Joris~M. Mooij.
\newblock Foundations of structural causal models with cycles and latent
  variables.
\newblock \emph{The Annals of Statistics}, 49\penalty0 (5):\penalty0
  2885--2915, 2021.

\bibitem[Dawid(2000)]{dawid00}
A.~Philip Dawid.
\newblock Causal inference without counterfactuals (with discussion).
\newblock \emph{Journal of the American Statistical Association}, 95:\penalty0
  465--480, 2000.

\bibitem[Dawid and Musio(2022)]{dawid22}
A.~Philip Dawid and Monica Musio.
\newblock Effects of causes and causes of effects.
\newblock \emph{Annual Review of Statistics and Its Application}, 9:\penalty0
  261--287, 2022.

\bibitem[Dawid and Senn(2023)]{dawid23}
A.~Philip Dawid and Stephen Senn.
\newblock Personalised decision-making without counterfactuals.
\newblock \emph{arXiv preprint}, https://arxiv.org/abs/2301.11976, 2023.

\bibitem[Galhotra and Halpern(2024)]{galhotra24}
Sainyam Galhotra and Joseph~Y. Halpern.
\newblock Intervention and conditioning in causal bayesian networks.
\newblock In \emph{Advances in Neural Information Processing Systems 37
  (NeurIPS 2024)}, volume~37, pages 89019--89041, 2024.

\bibitem[Galles and Pearl(1998)]{galles98}
David Galles and Judea Pearl.
\newblock An axiomatic characterization of causal counterfactuals.
\newblock \emph{Foundations of Science}, 3\penalty0 (1):\penalty0 151--182,
  1998.

\bibitem[Halpern(2000)]{halpern00}
Joseph~Y. Halpern.
\newblock Axiomatizing causal reasoning.
\newblock \emph{Journal of Artificial Intelligence Research}, 12:\penalty0
  317--337, 2000.

\bibitem[Halpern(2016)]{halpernbook}
Joseph~Y. Halpern.
\newblock \emph{Actual Causality}.
\newblock MIT Press, 2016.

\bibitem[Halpern and Peters(2022)]{halpern22}
Joseph~Y. Halpern and Spencer Peters.
\newblock Reasoning about causal models with infinitely many variables.
\newblock \emph{Proceedings of the 36th AAAI Conference on Artificial
  Intelligence}, 36\penalty0 (5668-5675), 2022.

\bibitem[Lewis(1973)]{lewis73a}
David Lewis.
\newblock \emph{Counterfactuals}.
\newblock Blackwell Publishers and Harvard University Press, 1973.

\bibitem[Mueller and Pearl(2023)]{mueller23}
Scott Mueller and Judea Pearl.
\newblock Personalized decision making -- a conceptual introduction.
\newblock \emph{Journal of Causal Inference}, 11\penalty0 (1), 2023.

\bibitem[Pearl(2009)]{pearl:book2}
Judea Pearl.
\newblock \emph{Causality: Models, Reasoning, and Inference; 2nd edition}.
\newblock Cambridge University Press, 2009.

\bibitem[Peters and Halpern(2021)]{peters21}
Spencer Peters and Joseph~Y. Halpern.
\newblock Causal modeling with infinitely many variables.
\newblock \emph{arXiv preprint}, https://arxiv.org/abs/2112.09171, 2021.

\bibitem[Robins(1987)]{robins1987}
James~M. Robins.
\newblock Addendum to ``a new approach to causal inference in mortality studies
  with a sustained exposure period---application to control of the healthy
  worker survivor effect''.
\newblock \emph{Computers \& Mathematics With Applications}, 14:\penalty0
  923--945, 1987.

\bibitem[Rubin(1974)]{rubin74}
Donald~B. Rubin.
\newblock Estimating causal effects of treatment in randomized and
  nonrandomized studies.
\newblock \emph{Journal of Educational Psychology}, 66:\penalty0 688--701,
  1974.

\bibitem[Sarvet and Stensrud(2023)]{sarvet23}
Aaron~L Sarvet and Mats~J Stensrud.
\newblock Perspective on `harm' in personalized medicine.
\newblock \emph{American Journal of Epidemiology}, 2023.

\bibitem[Spirtes et~al.(2001)Spirtes, Glymour, and Scheines]{spirtes00}
Peter Spirtes, Clark Glymour, and Richard Scheines.
\newblock \emph{Causation, Prediction, and Search}.
\newblock MIT Press, 2nd edition edition, 2001.

\bibitem[Tian and Pearl(2000)]{tian00}
Jin Tian and Judea Pearl.
\newblock Probabilities of causation: Bounds and identification.
\newblock \emph{Annals of Mathematics and Artificial Intelligence}, 28\penalty0
  (1):\penalty0 287--313, 2000.

\bibitem[Walters and Williams(2013)]{walters13}
Lee Walters and J.~Robert~G. Williams.
\newblock An argument for conjunction conditionalization.
\newblock \emph{The Review of Symbolic Logic}, 6\penalty0 (4):\penalty0
  573--588, 2013.

\bibitem[Wysocki(2023)]{wysocki24}
Tomasz Wysocki.
\newblock The underdeterministic framework.
\newblock \emph{British journal for the philosophy of science}, forthcoming,
  2023.

\bibitem[Zhang et~al.(2022)Zhang, Tian, and Bareinboim]{zhang22}
Junzhe Zhang, Jin Tian, and Elias Bareinboim.
\newblock Partial counterfactual identification from observational and
  experimental data.
\newblock In \emph{Proceedings of the 39th International Conference on Machine
  Learning}, volume 162, pages 26548--26558, 2022.

\end{thebibliography}

\appendix

\section{}

\subsection{Proofs of Theorems}

\begin{theorem*}{\bf \ref{thm:mod}} Given a nondeterministic causal model $M$, we have that for all $\vec{Y} \subseteq \V$, for all $\vec{y} \in \R(\vec{Y})$, and for all basic formulas $\phi$:
\begin{itemize}
\item $M \sat [\vec{Y} \gets \vec{y}]\phi$ iff $\vec{v} \sat \phi$ for all solutions $(\vec{u},\vec{v})$ of $M_{\vec{Y} \gets \vec{y}}$.
\item For all contexts $\vec{u}$: $(M,\vec{u}) \sat [\vec{Y} \gets \vec{y}]\phi$ iff $\vec{v} \sat \phi$ for all states $\vec{v}$ such that $(\vec{u},\vec{v})$ is a solution of $M_{\vec{Y} \gets \vec{y}}$.
\end{itemize}
\end{theorem*}

\prf
We first show that the two claims are equivalent. Per definition $M \sat [\vec{Y} \gets \vec{y}]\phi$ iff $(M,\vec{u},\vec{v}) \sat [\vec{Y} \gets \vec{y}]\phi$ for all solutions $(\vec{u},\vec{v})$ of $M$, which is equivalent to the statement that for all contexts $\vec{u}$: $(M,\vec{u},\vec{v}) \sat [\vec{Y} \gets \vec{y}]\phi$ for all states $\vec{v}$ such that $(\vec{u},\vec{v})$ is a solution of $M$. In turn, the latter is per definition equivalent to the statement that for all contexts $\vec{u}$: $(M,\vec{u}) \sat [\vec{Y} \gets \vec{y}]\phi$. By the second claim, the latter is equivalent to the statement that for all contexts $\vec{u}$: $\vec{v} \sat \phi$ for all states $\vec{v}$ such that $(\vec{u},\vec{v})$ is a solution of $M_{\vec{Y} \gets \vec{y}}$. Finally, this is equivalent to the right side of the first claim. 

We now prove the first claim. %Per definition, $M \sat [\vec{Y} \gets \vec{y}]\phi$ iff $(M,\vec{u},\vec{v}) \sat [\vec{Y} \gets \vec{y}]\phi$ holds for all solutions $(\vec{u},\vec{v})$ of $M$. 
Recall that, given a solution $(\vec{u},\vec{v})$ of $M$, $(M,\vec{u},\vec{v}) \sat [\vec{Y} \gets \vec{y}]\phi$ iff $\vec{v}' \sat \phi$ for all states $\vec{v}'$ such that $(\vec{u},\vec{v}')$ is a solution of $(M^{(\vec{u},\vec{v})})_{\vec{Y} \gets \vec{y}}$. Therefore we need to show that the following claims are equivalent:
\begin{itemize}
\item $\vec{v}' \sat \phi$ for all states $\vec{v}'$ for which there exists some solution $(\vec{u},\vec{v})$ of $M$ such that  $(\vec{u},\vec{v}')$ is a solution of $(M^{(\vec{u},\vec{v})})_{\vec{Y} \gets \vec{y}}$.
\item $\vec{v}' \sat \phi$ for all solutions $(\vec{u},\vec{v}')$ of $M_{\vec{Y} \gets \vec{y}}$. 
\end{itemize}

This equivalence follows directly from the statement that for all states $\vec{v}'$ and all contexts $\vec{u}$, the following claims are equivalent:
\begin{itemize}
\item $(\vec{u},\vec{v}')$ is a solution of $(M^{(\vec{u},\vec{v})})_{\vec{Y} \gets \vec{y}}$ for some solution $(\vec{u},\vec{v})$ of $M$.
\item $(\vec{u},\vec{v}')$ is a solution of $M_{\vec{Y} \gets \vec{y}}$.
\end{itemize}

The implication from the first claim to the second is a direct consequence of the definition of an actualized refinement. Therefore we proceed with proving the reverse implication.

Assume $(\vec{u},\vec{v}')$ is a solution of $M_{\vec{Y} \gets \vec{y}}$. We need to show that there exists some $\vec{v}$ such that $(\vec{u},\vec{v}')$ is a solution of $(M^{(\vec{u},\vec{v})})_{\vec{Y} \gets \vec{y}}$ and $(\vec{u},\vec{v})$ is a solution of $M$. We proceed by induction over the partial order given by $\G_M$, by considering the claim: given an ancestrally closed set $\vec{W} \subseteq \V$, there exists some $\vec{w}$ such that $(\vec{u},\vec{w}')$ appears in a solution of $(M^{(\vec{u},\vec{w})})_{\vec{Y} \gets \vec{y}}$ and the set of solutions extending $(\vec{u},\vec{w})$ of $M$ is non-empty. Here $\vec{w}' \subseteq \vec{v}'$, and $M^{(\vec{u},\vec{w})}$ instantiates the obvious generalization of Definition \ref{def:ar} to ancestrally closed partial settings.

Given that we only consider total models, there exists at least one solution (to both $M$ and to $M_{\vec{Y} \gets \vec{y}}$) for each context $\vec{u}$, and thus the claim holds for the base case with $\vec{W}=\emptyset$.

We now proceed with the inductive case. Assume that the set of solutions of $M$ extending $(\vec{u},\vec{w})$ is non-empty and $(\vec{u},\vec{w}')$ appears in some solution of $(M^{(\vec{u},\vec{w})})_{\vec{Y} \gets \vec{y}}$. 

Say $V$ is the next variable in $\V - \vec{W}$ according to the partial order of $\G_M$, and $v' \in \vec{v}'$. We need to prove that there exists some $v \in \R(V)$ so that the set of solutions of $M$ extending $(\vec{u},\vec{w},v)$ is non-empty and such that $(\vec{u},\vec{w}',v')$ appears in some solution of $(M^{(\vec{u},\vec{w},v)})_{\vec{Y} \gets \vec{y}}$. 

First we consider the case where $V \in \vec{Y}$. Since the set of solutions extending $(\vec{u},\vec{w})$ is non-empty, there exists some $v$ such that the set of solutions extending $(\vec{u},\vec{w},v)$ is non-empty as well. The equation for $V$ in $(M^{(\vec{u},\vec{w},v)})_{\vec{Y} \gets \vec{y}}$ will be the constant equation $V=v^*$, where $v^* \in \vec{y}$. Furthermore, since $(\vec{u},\vec{v}')$ is a solution of $M_{\vec{Y} \gets \vec{y}}$, $v^*=v'$. It follows that $(\vec{u},\vec{w}',v')$ appears in a solution of $(M^{(\vec{u},\vec{w},v)})_{\vec{Y} \gets \vec{y}}$, as had to be shown.

Second we consider the case where $V \not \in \vec{Y}$, which we separate into two sub-cases. Here $\vec{pa_V}' \subseteq \vec{v}'$. Note that, as $\vec{W}$ is ancestrally closed, $\vec{Pa_V} \subseteq \vec{W}$.

Consider the case where $\vec{pa_V} \neq \vec{pa_V}'$, where $\vec{pa_V} \subseteq \vec{w}$. As before, we can consider any $v$ such that the set of solutions extending $(\vec{u},\vec{w},v)$ is non-empty. Per definition of an actualized refinement, we have that $f_V(\vec{pa_V}')=f^{(v,\vec{pa_V})}_V(\vec{pa_V}')$. Furthermore, since $(\vec{u},\vec{v}')$ is a solution of $M_{\vec{Y} \gets \vec{y}}$, we have that $v' \in f_V(\vec{pa_V}')$, and thus also $v' \in f^{(v,\vec{pa_V})}_V(\vec{pa_V}')$. It follows that $(\vec{u},\vec{w}',v')$ appears in a solution of $(M^{(\vec{u},\vec{w},v)})_{\vec{Y} \gets \vec{y}}$.

Lastly, consider the case where $\vec{pa_V} = \vec{pa_V}'$. As each $v \in f_V(\vec{pa_V})$ is a solution of $V=f_V(\vec{Pa_X})$, each solution of $M$ that starts with $(\vec{u},\vec{w})$ can be extended to a solution $(\vec{u},\vec{w},v)$ for each $v \in f_V(\vec{pa_V})$. Therefore the set of solutions extending $(\vec{u},\vec{w},v)$ is non-empty for each $v \in f_V(\vec{pa_V})$. Choosing $v = v'$, we get that there exists at least one solution that extends $(\vec{u},\vec{w},v)$, and $(\vec{u},\vec{w}',v')$ appears in a solution of $(M^{(\vec{u},\vec{w},v)})_{\vec{Y} \gets \vec{y}}$. This concludes the proof.
\eprf

\begin{theorem*}{\bf \ref{thm:axiom}} $AX^{swc}_{non}$ (resp. $AX^{scc}_{non}$) is a sound and complete axiomatization for the language $\cal{L}(\S)$ with respect to the {\bf swc} logic (resp. the {\bf scc} logic) over acyclic NSCMs that have signature $\S$.
\end{theorem*}

\prf We start with completeness. First we consider $AX^{swc}_{non}$ and the {\bf swc} logic. It suffices to show that for any formula $\phi \in \cal{L}(\S)$ that is consistent with $AX^{swc}_{non}(\S)$, there is an acyclic NSCM $M$ such that $(M,\vec{u},\vec{v}) \sat \phi$ for some causal setting $(M,\vec{u},\vec{v})$. The proof follows the same technique as used by \cite{halpern00} and \cite{halpern22}. 

%Note that an acyclic NSCM is a universal finite acyclic GSEM!
%AX+basic: D0,1,2,4,7,8

Suppose that $\phi \in \cal{L}(\S)$ is consistent with $AX^{swc}_{non}(\S)$ (i.e., we cannot prove $\lnot \phi$ in $AX^{swc}_{non}(\S)$). Then $\phi$ can be extended to a maximal consistent set $C$ of formulas, meaning that $\phi \in C$, every finite subset $C'$ of $C$ is consistent with $AX^{swc}_{non}(\S)$, and no strict superset $C^*$ of $C$ has the property that every finite subset of $C^*$ is consistent with $AX^{swc}_{non}(\S)$. Standard arguments show that, for every formula $\psi \in \cal{L}(\S)$, either $\psi$ or $\lnot \psi$ must be in $C$. Moreover, every instance of the axioms in $AX^{swc}_{non}(\S)$ must be in $C$. 

We now define an acyclic NSCM $M$ with signature $\S$ and the requisite causal setting $(M,\vec{u},\vec{v})$ as follows. Fix some context $\vec{u} \in \R(\U)$. The following Lemma from \cite{halpern22} is useful, where $AX_{bas}$ is the axiom system consisting of D0, D7, D8, and MP.

\lem\label{lem:lem1} 

(a) $AX_{bas} \vdash [\vec{Y} \gets \vec{y}]\phi_1 \land [\vec{Y} \gets \vec{y}]\phi_2 \Leftrightarrow [\vec{Y} \gets \vec{y}](\phi_1 \land \phi_2)$

(b) $AX_{bas} \vdash \langle\vec{Y} \gets \vec{y}\rangle(\phi_1 \lor \phi_2) \Leftrightarrow \langle\vec{Y} \gets \vec{y}\rangle\phi_1 \lor \langle\vec{Y} \gets \vec{y}\rangle\phi_2$
\elem

Applying Lemma \ref{lem:lem1} and the axioms D1-2, D7, and D10(a) to the empty intervention, it follows by standard modal reasoning that there exists some $\vec{v} \in \R(\V)$ so that  $\langle \rangle \vec{V}=\vec{v} \in C$. We will construct $M$ such that $(M,\vec{u},\vec{v}) \sat \phi$. 

For all $X \in \V$ we define $\vec{Pa_X}$ as the set of variables $Y \in \V$ such that $Y \rightsquigarrow X \in C$. Given D6, we have hereby defined an acyclic graph $\G$. We define $F_X$ for $x \in \R(X)$ and $\vec{pa_X} \in \R(\vec{Pa_X})$ by taking $x \in f_X(\vec{pa_X})$ iff $\langle\vec{Pa_X} \gets \vec{pa_x} \rangle X=x \in C$. As before, it follows from Lemma \ref{lem:lem1} and the axioms D1-2, D7, and D10(a), that for each choice $\vec{pa_X}$ there will be at least one value $x$ such that $x \in f_X(\vec{pa_X})$, and thus our NSCM is total. Therefore $M$ is an acyclic and total NSCM, as required. 

Furthermore, $\langle \rangle \vec{V}=\vec{v} \in C$ combined with D10(c) gives $[]\vec{V}=\vec{v} \in C$. Applying D3(b) and Lemma  \ref{lem:lem1}, for each $X$ we get that $[\vec{Pa_X} \gets \vec{pa_X}]X=x \in C$, where $(x,\vec{pa_X}) \subseteq \vec{v}$. Therefore the function $f_X$ is deterministic for the actual parent values, (i.e, $x = f_X(\vec{pa_X})$), and thus taking $f^{(\vec{pa_X},x)}_X(\vec{pa_X})=f_X(\vec{pa_X})$ results in a deterministic actualized refinement for the actual parent values, as required for an NSCM. It remains to be shown that $(M,\vec{u},\vec{v}) \sat \phi$. 

Following exactly the same reasoning as in the proof of Theorem 5.2 in \cite{halpern22} and in \cite{beckers23a}, it follows that this reduces to showing for all formulas of the form $\langle\vec{Y} \gets \vec{y}\rangle\vec{X}=\vec{x}$ with $\vec{X}=\V - \vec{Y}$ that $\langle\vec{Y} \gets \vec{y}\rangle\vec{X}=\vec{x} \in C$ iff $(M,\vec{u},\vec{v}) \sat \langle\vec{Y} \gets \vec{y}\rangle\vec{X}=\vec{x}$.

Suppose that $\langle\vec{Y} \gets \vec{y}\rangle \vec{X}=\vec{x} \in C$. Let $\vec{v'}=(\vec{y},\vec{x})$. It suffices to show that for each $X \in \vec{X}$, $\langle\vec{Pa_X} \gets \vec{pa_x} \rangle X=x \in C$, where $(x,\vec{pa_X}) \subseteq \vec{v'}$. Consider some $X$ and the requisite values $\vec{pa_X}$ and $x$. By D3(a), we have that $\langle\vec{Pa_X} \gets \vec{pa_X},\vec{Z} \gets \vec{z} \rangle X=x \in C$, where $\vec{Z}= \V - (\vec{Pa_X} \cup \{X\})$, and $\vec{z} \subseteq \vec{v'}$.

Let us consider some $\vec{z''} \in \R(\vec{Z})$ such that $\langle\vec{Pa_X} \gets \vec{pa_X}, X \gets x\rangle (\vec{Z} = \vec{z''}) \in C$. (As before, such $z'' \in \R(Z)$ must exist.) If also $\langle\vec{Pa_X} \gets \vec{pa_X}, \vec{Z} \gets \vec{z''}\rangle X = x \in C$, then by D5 and D7 we get that $\langle\vec{Pa_X} \gets \vec{pa_X}\rangle X = x \in C$, as required. 
			      
Remains to consider the case where $[\vec{Pa_X} \gets \vec{pa_X}, \vec{Z} \gets \vec{z''}] X \neq x \in C$. It follows that $\vec{z} \neq \vec{z''}$. Per construction of $\vec{Pa_X}$, we have for each $Z \in \vec{Z}$ that $\lnot (Z \rightsquigarrow X) \in C$. We show by induction that this results in a contradiction. 

For the base case, take $\vec{A}_0=\emptyset$, and $\vec{a}_0 \subseteq \vec{z}$. Let $\vec{W}_0 = \vec{Z} - \vec{A}_0$, and $\vec{w}_0 \subseteq \vec{z}$, $\vec{w''}_0 \subseteq \vec{z''}$. We have that $[\vec{Pa_X} \gets \vec{pa_X}, \vec{W}_0 \gets \vec{w''}_0,\vec{A}_0 \gets \vec{a}_0] X \neq x \in C$. 

The inductive case consists of considering $\vec{A}_{k+1}=\vec{A}_k \cup \{Z\}$ for some $Z \in \vec{W}_k$. We let $\vec{W}_{k+1}= \vec{Z} - \vec{A}_{k+1}$, $\vec{a}_{k+1} \subseteq \vec{z}$, and $\vec{w}_{k+1} \subseteq \vec{z}$, $\vec{w''}_{k+1} \subseteq \vec{z''}$. By the induction hypothesis, we have that $[\vec{Pa_X} \gets \vec{pa_X}, \vec{W}_k \gets \vec{w''}_k,\vec{A}_k \gets \vec{a}_k] X \neq x \in C$, which can be rewritten as $[\vec{Pa_X} \gets \vec{pa_X}, \vec{W}_{k+1} \gets \vec{w''}_{k+1},\vec{A}_k \gets \vec{a}_k,Z \gets z''] X \neq x \in C$. If $\langle\vec{Pa_X} \gets \vec{pa_X}, \vec{W}_{k+1} \gets \vec{w''}_{k+1}, \vec{A}_k \gets \vec{a}_k, Z \gets z\rangle X =x \in C$, it follows that $Z \rightsquigarrow X \in C$. Therefore, $[\vec{Pa_X} \gets \vec{pa_X}, \vec{W}_{k+1} \gets \vec{w''}_{k+1}, \vec{A}_{k+1} \gets \vec{a}_{k+1}] X \neq x \in C$. Given that $| \vec{Z} |$ is finite, for some $k \in \mathbb{N}$ this results in a contradiction.

For the other way, suppose that $(M,\vec{u},\vec{v}) \sat \langle\vec{Y} \gets \vec{y}\rangle\vec{X}=\vec{x}$. Let $\vec{v'}=(\vec{y},\vec{x})$. Per construction of $M$, we know that for each $X \in \vec{X}$, $\langle\vec{Pa_X} \gets \vec{pa_x} \rangle X=x \in C$, where $(x,\vec{pa_X}) \subseteq\vec{v'})$.

For each $X \in \vec{X}$, we have that for any $\vec{Z} \subseteq \V - (\vec{Pa_X} \cup \{X\})$: $\lnot (Z_i \rightsquigarrow X) \in C$ for all $Z_i \in \vec{Z}$. Therefore, for any $\vec{z}$ we have that $\langle\vec{Z} \gets \vec{z}, \vec{Pa_X} \gets \vec{pa_x} \rangle X=x \in C$. Letting $\vec{X} = \{X_1,\ldots,X_k\}$, we have in particular that for each $i \in \{1,\ldots,k\}$: $\langle\vec{Y} \gets \vec{y}, \vec{X}^{-i} \gets \vec{x}^{-i} \rangle X_i=x_i \in C$, where $\vec{X}^{-i} := (X_1, \ldots, X_{i-1},X_{i+1},\ldots,X_k)$.

Taking $\langle\vec{Y} \gets \vec{y}, \vec{X}^{-1} \gets \vec{x}^{-1}  \rangle X_1=x_1 \in C$ and $\langle\vec{Y} \gets \vec{y}, \vec{X}^{-2} \gets \vec{x}^{-2}  \rangle X_2=x_2 \in C$, we can apply D5 to derive that  $\langle\vec{Y} \gets \vec{y}, \vec{X}^{-1,2} \gets \vec{x}^{-1,2} \rangle (X_1=x_1 \land X_2=x_2)\in C$. By the same reasoning, we get that $\langle\vec{Y} \gets \vec{y}, \vec{X}^{-2,3} \gets \vec{x}^{-2,3} \rangle (X_2=x_2 \land X_3=x_3)\in C$. Again applying D5 to the last two statements, we get that  $\langle\vec{Y} \gets \vec{y}, \vec{X}^{-1,2,3} \gets \vec{x}^{-1,2,3} \rangle (X_1=x_1 \land X_2=x_2 \land X_3=x_3)\in C$. By straightforward induction, we get that $\langle\vec{Y} \gets \vec{y} \rangle \vec{X} \gets \vec{x} \in C$, which is what had to be shown. This concludes the proof of completeness.

For $AX^{scc}_{non}$ and the {\bf scc} logic, the proof proceeds identically except for three differences. The first difference is that we have no need for some $\vec{v} \in \R(\V)$ so that  $\langle \rangle \vec{V}=\vec{v} \in C$, as we only need to construct $M$ and $\vec{u}$ such that $(M,\vec{u}) \sat \phi$. The second (related) difference is that in this case there is no need to consider some actual values $\vec{pa_X}$ for each $X \in \vec{X}$ and verify that $f_X(\vec{pa_X})$ is deterministic, for the actualized refinement is not relevant to the semantics of  $\sat$ for $(M,\vec{u})$, as shown by Theorem \ref{thm:mod}. This explains why D10(c) is not part of $AX^{scc}_{non}$. The third difference is that we here need to show $\langle\vec{Y} \gets \vec{y}\rangle\vec{X}=\vec{x} \in C$ iff $(M,\vec{u}) \sat \langle\vec{Y} \gets \vec{y}\rangle\vec{X}=\vec{x}$ (as opposed to having $\vec{v}$ included on the RHS). By Theorem \ref{thm:mod}, this is equivalent to showing that  $\langle\vec{Y} \gets \vec{y}\rangle\vec{X}=\vec{x} \in C$ iff $(\vec{u},\vec{y},\vec{x})$ is a solution of $M_{\vec{Y} \gets \vec{y}}$. The remainder of the proof remains identical.

Now we prove soundness. We leave it as a simple exercise to the reader that D0-D1-D2-D4-D7-D8-D10(a) are sound for both of our logics, and that D10(c) is sound for our {\bf swc} logic.

{\bf D6}: 

As we explain in the discussion of \cite{wysocki24}'s work in Section \ref{sec:rel}, contrary to our NSCMs, DSCMs do not come with a graph. Rather, a graph $\G_D$ is induced by invoking ``$Y$ depends on $X$'': there is an edge from $X$ to $Y$ iff there exist settings $\vec{z} \in (\U \cup \V - \{X,Y\}$, and $x,x' \in \R(X)$, such that $f_Y(\vec{z},x) \neq f_Y(\vec{z},x')$. As mentioned in \cite{halpern22}, D6 expresses the acyclicity of the induced graph $\G_D$. Of course NSCMs can be used to invoke a graph $\G_D$ in exactly the same manner: there is an edge from $X$ to $Y$ iff there exist settings  $\vec{z} \in (\U \cup \V - \{X,Y\}$, and $x,x' \in \R(X)$, such that $f_Y(\vec{z},x) \neq f_Y(\vec{z},x')$. NSCMs also come with an explicit graph $\G_M$, and this need not be identical to $\G_D$. But the graph $\G_D$ is easily seen to be a subgraph of $\G_M$, since $X$ has to be an argument of $f_Y$ for $Y$ to depend on $X$, and per definition this means that there is an edge from $X$ to $Y$ in $\G_M$. Therefore the acyclicity of $\G_M$ implies the acyclicity of $\G_D$, and thus D6 is sound for both of our logics. Concretely: if $X \rightsquigarrow Y$ for some solution $(\vec{u},\vec{v})$, then $X$ has to be an ancestor of $Y$ in $\G_D$. So the falsity of D6 would imply that $\G_D$ is cyclic.
 
{\bf D3(a)}:

We start with soundness for $\sat$ relative to $(M,\vec{u},\vec{v})$. Assume $W \not \in \vec{X}$, and $(\vec{u},\vec{v})$ is a solution of $M$, and $(M,\vec{u},\vec{v}) \sat \langle\vec{X} \gets \vec{x}\rangle(W =w  \land \phi)$. This means that there exists $\vec{v}'$ so that  $(\vec{u},\vec{v}')$ is a solution of $(M^{(\vec{u},\vec{v})})_{\vec{X} \gets \vec{x}}$ and $\vec{v'} \sat (W =w \land \phi)$. By D4, it directly follows that $(\vec{u},\vec{v}')$ is also a solution of $(M^{(\vec{u},\vec{v})})_{\vec{X} \gets \vec{x}, W \gets w}$, and we also have that  $\vec{v'} \sat \phi$. This means precisely that $(M,\vec{u},\vec{v}) \sat \langle\vec{X} \gets \vec{x}, W \gets w\rangle\phi$.

Soundness for $\sat$ relative to $(M,\vec{u})$ is a consequence of the soundness for $\sat$ relative to $(M,\vec{u},\vec{v})$. For concreteness, we here write out the intermediate steps. Assume that $(M,\vec{u}) \sat \langle\vec{X} \gets \vec{x}\rangle(W =w  \land \phi)$. Per definition of $\langle\rangle$, this is equivalent to: $(M,\vec{u}) \sat \lnot [\vec{X} \gets \vec{x}](W \neq w  \lor \lnot \phi)$. In turn, this is equivalent to it not being the case that for all states $\vec{v}$ so that $(\vec{u},\vec{v})$ is a solution of $M$, we have $(M,\vec{u},\vec{v}) \sat  [\vec{X} \gets \vec{x}](W \neq w  \lor \lnot \phi)$. This is equivalent to there existing some state $\vec{v'}$ such that $(\vec{u},\vec{v'})$ is a solution of $M$ and $(M,\vec{u},\vec{v'}) \sat  \langle\vec{X} \gets \vec{x}\rangle(W =w  \land \phi)$. By D3(a) for $\sat$ relative to causal settings, we get that for some solution $(\vec{u},\vec{v''})$ of $M$, namely $(\vec{u},\vec{v'})$, $(M,\vec{u},\vec{v''}) \sat  \langle\vec{X} \gets \vec{x}, W \gets w\rangle \phi$. Applying all of the above equivalences in the other direction, this is seen to be equivalent to $(M,\vec{u}) \sat  \langle\vec{X} \gets \vec{x}, W \gets w\rangle \phi$, which is what had to be shown.

{\bf D3(b)}: 

Soundness for $\sat$ relative to $(M,\vec{u})$ is a direct consequence of the soundness for $\sat$ relative to $(M,\vec{u},\vec{v})$, so we proceed with the latter.

Assume $W \not \in \vec{X}$, and $(\vec{u},\vec{v})$ is a solution of $M$, and $(M,\vec{u},\vec{v}) \sat [\vec{X} \gets \vec{x}](W =w  \land \phi)$. It suffices to show that for any state $\vec{v}'$, if $(\vec{u},\vec{v}')$ is a solution of $(M^{(\vec{u},\vec{v})})_{\vec{X} \gets \vec{x}, W \gets w}$, then it is a solution of $(M^{(\vec{u},\vec{v})})_{\vec{X} \gets \vec{x}}$.

We proceed by a reductio. Assume that $(\vec{u},\vec{v}')$ is a solution only of the former. Note that the two models have identical equations for all variables except $W$, and that it has to be that $w \in \vec{v}'$. Therefore it must be that $(w,\vec{pa_W})$ is not a solution of $W$'s equation in $(M^{(\vec{u},\vec{v})})_{\vec{X} \gets \vec{x}}$, where $\vec{pa_W} \subseteq \vec{v}')$. Given that, per the first assumption, $W=w$ for all solutions of  $(M^{(\vec{u},\vec{v})})_{\vec{X} \gets \vec{x}}$, it must be that $\vec{pa_W}$ does not appear in any such solution. However, given that both models are acyclic and have identical equations for all non-descendants of $W$ (including $\vec{pa_W}$), they have exactly the same partial solutions $(\vec{u},\vec{a})$, where $\vec{A} \subseteq \V$ consists of all non-descendants of $W$. Thus it cannot be that $\vec{pa_W}$ only appears in a solution to one of them.

{\bf D5}:

We start with soundness for $\sat$ relative to $(M,\vec{u},\vec{v})$. Assume $(\vec{u},\vec{v})$ is a solution of $M$, and  $(M,\vec{u},\vec{v}) \sat \<\XX \gets \xx, Y \gets y\> (W = w \land \ZZ = \zz)  \land   \<\XX \gets \xx, W \gets w\> (Y = y \land \ZZ = \zz)$, where $\vec{Z} = \V - (\vec{X} \cup \{W,Y\})$. We need to show that $(M,\vec{u},\vec{v}) \sat \<\XX \gets \xx\> (W = w \land Y = y \land \ZZ = \zz)$. 

Per assumption, there exists a solution $(\vec{u},\vec{v_1})$ of $(M^{(\vec{u},\vec{v})})_{\vec{X} \gets \vec{x}, Y \gets y}$ such that its restriction to $(W,\vec{Z})$ is $(w,\vec{z})$, and there exists a solution $(\vec{u},\vec{v_2})$ of $(M^{(\vec{u},\vec{v})})_{\vec{X} \gets \vec{x}, W \gets w}$ such that its restriction to $(Y,\vec{Z})$ is $(y,\vec{z})$. Clearly also both solutions have that $\vec{X}=\vec{x}$, and it must be that $y \in \vec{v_1}$, and also $w \in \vec{v_2}$.

Since $\V=\vec{Z} \cup \vec{X} \cup \{W,Y\}$, the tuple $\vec{v_3}=(\vec{x},\vec{z},y,w)$ is a state. Therefore it follows that $\vec{v_1}=\vec{v_2}=\vec{v_3}$. It now suffices to show that $(\vec{u},\vec{v_3})$ is a solution of $(M^{(\vec{u},\vec{v})})_{\vec{X} \gets \vec{x}}$. As the equations for all variables in $\vec{Z} \cup \vec{X}$ are identical across the three models, $(\vec{u},\vec{v_3})$ contains solutions to all these equations in $(M^{(\vec{u},\vec{v})})_{\vec{X} \gets \vec{x}}$. As the equation for $Y$ is identical across $(M^{(\vec{u},\vec{v})})_{\vec{X} \gets \vec{x}}$ and $(M^{(\vec{u},\vec{v})})_{\vec{X} \gets \vec{x}, W \gets w}$, $(\vec{u},\vec{v_3})$ contains a solution for $Y$ in $(M^{(\vec{u},\vec{v})})_{\vec{X} \gets \vec{x}}$. The same holds for $W$ and the other model, concluding the proof.

Now we prove soundness for $\sat$ relative to $(M,\vec{u})$. Assume that $(M,\vec{u}) \sat \<\XX \gets \xx, Y \gets y\> (W = w \land \ZZ = \zz)  \land   \<\XX \gets \xx, W \gets w\> (Y = y \land \ZZ = \zz)$, where $\vec{Z} = \V - (\vec{X} \cup \{W,Y\})$. We need to show that $(M,\vec{u}) \sat \<\XX \gets \xx\> (W = w \land Y = y \land \ZZ = \zz)$. By Theorem \ref{thm:mod}, our assumption is equivalent to the statement that $(\vec{u},\vec{x},\vec{z},y,w)$ is a solution of both $M_{\vec{X} \gets \vec{x}, Y \gets y}$ and $M_{\vec{X} \gets \vec{x}, W \gets w}$, and what we need to show is that $(\vec{u},\vec{x},\vec{z},y,w)$ is a solution of $M_{\vec{X} \gets \vec{x}}$. This follows by applying the  reasoning from the previous paragraph to the three models at hand. 
\eprf

\begin{proposition*}{\bf \ref{pro:act}}
If $(M,\vec{u},\vec{v}) \sat \vec{X} =\vec{x}  \land \phi $ then $(M,\vec{u},\vec{v}) \sat [\vec{X} \gets \vec{x}] \phi$.
\end{proposition*}

\prf 
We show that for any $\phi'$: $(M,\vec{u},\vec{v}) \sat\phi' $ iff $(M,\vec{u},\vec{v}) \sat []\phi' $. The result then follows by repeated application of D3(b), starting with the antecedent $[](\vec{X} =\vec{x}  \land \phi)$. 

$(M,\vec{u},\vec{v}) \sat \phi'$ simply means that $\vec{v} \sat \phi'$. So it suffices to show that $(\vec{u},\vec{v})$ is the only solution of $M^{(\vec{u},\vec{v})}$ that extends the context $\vec{u}$. We proceed by induction by extending $\vec{u}$ one variable at the time, where we respect the partial order determined by $\G$. For the base case, given that exogenous variables have no equations, we trivially have that $\vec{u}$ is part of any solution extending $\vec{u}$. For the inductive case, assume that we have solved the equations for some set of variables $\vec{W} \subseteq \V$ and using $\U=\vec{u}$, and that this resulted in a unique solution  $(\vec{u},\vec{w})$ such that $\vec{w}$ agrees with $\vec{v}$ on $\vec{W}$. Consider some $X \in \V$ such that $\vec{Pa_X} \subseteq \vec{W} \cup \U$ (which has to exist giving acyclicity), and let $\vec{pa_X}$ be the restriction of $(\vec{u},\vec{w})$ to $\vec{Pa_X}$. Let $x \in \vec{v}$. Per definition of an actualized refinement, $f^{(\vec{pa_X},x)}_X(\vec{pa_X})=x$, and thus $(\vec{u},\vec{w},x)$ is the unique solution to the equations for the variables $\vec{W} \cup \{X\}$, which is what had to be shown.
\eprf

\begin{theorem*}{\bf \ref{thm:gen}} Say we have a PNSEM $M$ and a NSEM $M^*$ over the same signature and graph which are {\em consistent} with each other, meaning that for all $X \in \V$ and all $x \in \R(X)$, $\vec{pa_X} \in \R(\vec{Pa_X})$: $P_X(x | \vec{pa_X}) > 0$ iff $x \in f_X(\vec{pa_X})$. Then for
 all worlds $(\vec{u},\vec{v})$ and all $\vec{y}_{\vec{x}}, \ldots, \vec{z}_{\vec{w}}$ for some $\vec{Y},\vec{X}, \ldots, \vec{Z},\vec{W} \subseteq \V$:
\begin{align*}
P_M(\vec{y}_{\vec{x}},\ldots, \vec{z}_{\vec{w}} | \vec{u},\vec{v})=1
%$P_M(\vec{y}_{\vec{x}} | \vec{u},\vec{v}) \ldots P_M(\vec{z}_{\vec{w}} | \vec{u}, \vec{v})=1$
\text{ iff }(M^*,\vec{u},\vec{v}) \sat [\vec{X} \gets \vec{x}]\vec{Y} =\vec{y} \land \ldots \land [\vec{W} \gets \vec{w}]\vec{Z}=\vec{z}.
\end{align*}
\end{theorem*}

\prf
Note that we have (1):

$P_M(\vec{y}_{\vec{x}},\ldots, \vec{z}_{\vec{w}} | \vec{u},\vec{v})=1$ iff $P_M(\vec{y}_{\vec{x}} | \vec{u},\vec{v}) \ldots P_M(\vec{z}_{\vec{w}} | \vec{u},\vec{v})=1$

iff $P_M(\vec{y}_{\vec{x}} | \vec{u},\vec{v})=1$ and $\ldots$ and $P_M(\vec{z}_{\vec{w}} | \vec{u},\vec{v})=1$.

And also (2):

$(M^*,\vec{u},\vec{v}) \sat [\vec{X} \gets \vec{x}]\vec{Y} =\vec{y} \land \ldots \land [\vec{W} \gets \vec{w}]\vec{Z}=\vec{z}$ 

iff $(M^*,\vec{u},\vec{v}) \sat [\vec{X} \gets \vec{x}]\vec{Y} =\vec{y}$ and $\ldots$ and $(M^*,\vec{u},\vec{v}) \sat [\vec{W} \gets \vec{w}]\vec{Z}=\vec{z}$. 

Therefore it suffices to show that $P_M(\vec{y}_{\vec{x}} | \vec{u},\vec{v})=1$ iff $(M^*,\vec{u},\vec{v}) \sat [\vec{X} \gets \vec{x}]\vec{Y} =\vec{y}$, which is a direct consequence of Lemma \ref{lem:con}
\eprf

\lem\label{lem:con} Given a PNSEM $M$ and a NSEM $M^*$ that are {\em consistent} with each other, for
 all pair of worlds $(\vec{u},\vec{v})$, $(\vec{u},\vec{v}^*)$, and all $\vec{y}_{\vec{x}}$ for some $\vec{Y},\vec{X} \subseteq \V$:

$(\vec{u},\vec{v}^*)$ is a solution of $(M^{(\vec{u}, \vec{v})})_{\vec{X} \gets \vec{x}}$ iff $(\vec{u},\vec{v}^*)$ is a solution of $(M^{*(\vec{u}, \vec{v})})_{\vec{X} \gets \vec{x}}$.
\elem

\prf $(\vec{u},\vec{v}^*)$ is a solution of $(M^{*(\vec{u}, \vec{v})})_{\vec{X} \gets \vec{x}}$ 

\begin{centering}

 iff 
 
 \end{centering}
\begin{itemize}
\item $\vec{x} \subseteq \vec{v}^*$ and 
\item for all $Y \in \V \setminus \vec{X}$: $y^* \in f_Y^{(\vec{pa_Y},y)}(\vec{pa_Y}^*)$, where $(y,\vec{pa_Y}) \subseteq (\vec{u},\vec{v})$ and $(y^*,\vec{pa_Y}^*) \subseteq (\vec{u},\vec{v}^*)$.
\end{itemize}

\begin{centering}

 iff 
 
 \end{centering}

\begin{itemize}
\item  $\vec{x} \subseteq \vec{v}^*$ and 
\item for all $Y \in \V \setminus \vec{X}$ such that $\vec{pa_Y} = \vec{pa_Y}^*$: $y^*=y$
\item for all $Y \in \V \setminus \vec{X}$ such that $\vec{pa_Y} \neq \vec{pa_Y}^*$: 
 $y^* \in f_Y(\vec{pa_Y^*})$
\end{itemize}

\begin{centering}

 iff 
 
 \end{centering}
 
 \begin{itemize}
\item $\vec{x} \subseteq \vec{v}^*$ and 
\item for all $Y \in \V \setminus \vec{X}$ such that $\vec{pa_Y} = \vec{pa_Y}^*$: $y^*=y$
\item for all $Y \in \V \setminus \vec{X}$ such that $\vec{pa_Y} \neq \vec{pa_Y}^*$: 
 $y^* \in f_Y(\vec{pa_Y^*})$
\end{itemize}

\begin{centering}

 iff 
 
 \end{centering}

\begin{itemize}
\item $\vec{x} \subseteq \vec{v}^*$ and 
\item for all $Y \in \V \setminus \vec{X}$ such that $\vec{epa_Y} = \vec{epa_Y}^*$: $y^*=y$
\item for all $Y \in \V \setminus \vec{X}$ such that $\vec{epa_Y} \neq \vec{epa_Y}^*$:  $P_Y(y^* | \vec{epa_Y^*},\vec{xpa_Y}) > 0$
\end{itemize}

\begin{centering}

 iff 
 
 \end{centering}

$P_{(M^{(\vec{u}, \vec{v})})_{\vec{X} \gets \vec{x}}}(\vec{u},\vec{v}^*) > 0$

\begin{centering}

 iff 
 
 \end{centering}

$(\vec{u},\vec{v}^*)$ is a solution of $(M^{(\vec{u}, \vec{v})})_{\vec{X} \gets \vec{x}}$.

\eprf 

\begin{theorem*}{\bf \ref{thm:pdscm}}
Given a PDSCM $M$ and $\vec{Y},\vec{X}, \ldots, \vec{Z},\vec{W} \subseteq \V$, it holds that
\begin{align*}
P_M(\vec{y}_{\vec{x}}, \ldots, \vec{z}_{\vec{w}})= \sum_{\{\vec{u} | (M,\vec{u}) \sat [\vec{X} \gets \vec{x}]\vec{Y} =\vec{y} \land \ldots \land [\vec{W} \gets \vec{w}]\vec{Z}=\vec{z}\}}
P_M(\vec{u}).
\end{align*}
\end{theorem*}

\prf
From the fact that all $P_X(X | \vec{Pa_X})$ only take value in $\{0,1\}$ it follows directly that also $P_M(\vec{y}_{\vec{x}}, \ldots, \vec{z}_{\vec{w}} | \vec{u},\vec{v}) \in \{0,1\}$ for any  $\vec{y}_{\vec{x}}, \ldots, \vec{z}_{\vec{w}}$. 

Furthermore, note that a PDSCM $M$ can be viewed equivalently as the combination of a DSCM that is consistent with it together with $P_M(\U)$. Applying Theorem \ref{thm:gen}, this gives for any $(\vec{u},\vec{v})$ that: 

$P_M(\vec{y}_{\vec{x}}, \ldots, \vec{z}_{\vec{w}} | \vec{u},\vec{v}) =1$ iff $(M,\vec{u},\vec{v}) \sat [\vec{X} \gets \vec{x}]\vec{Y} =\vec{y} \land \ldots \land [\vec{W} \gets \vec{w}]\vec{Z}=\vec{z}$. 

Also, from the discussion in Section \ref{sec:com} we know that when our semantics is applied to a DSCM it reduces to the standard one, so that we get 

$(M,\vec{u},\vec{v}) \sat [\vec{X} \gets \vec{x}]\vec{Y} =\vec{y} \land \ldots \land [\vec{W} \gets \vec{w}]\vec{Z}=\vec{z}$ iff $(\vec{u},\vec{v})$ is a solution of $M$ and $(M,\vec{u}) \sat [\vec{X} \gets \vec{x}]\vec{Y} =\vec{y} \land \ldots \land [\vec{W} \gets \vec{w}]\vec{Z}=\vec{z}$ iff $(M,\vec{u}) \sat \V=\vec{v}$ and $(M,\vec{u}) \sat [\vec{X} \gets \vec{x}]\vec{Y} =\vec{y} \land \ldots \land [\vec{W} \gets \vec{w}]\vec{Z}=\vec{z}$.

Combining all three observations, we get that: 

$P_M(\vec{y}_{\vec{x}}, \ldots, \vec{z}_{\vec{w}})=\sum_{\{\vec{u},\vec{v}\}} P_M(\vec{y}_{\vec{x}}, \ldots, \vec{z}_{\vec{w}} | \vec{u},\vec{v}) P_M(\vec{u},\vec{v}) \\
=\sum_{\{\vec{u},\vec{v} | (M,\vec{u},\vec{v}) \sat [\vec{X} \gets \vec{x}]\vec{Y} =\vec{y} \land \ldots \land [\vec{W} \gets \vec{w}]\vec{Z}=\vec{z}\}}
P_M(\vec{u},\vec{v})\\
=\sum_{\{\vec{u},\vec{v} | (M,\vec{u}) \sat \V =\vec{v} \text{ and } (M,\vec{u}) \sat [\vec{X} \gets \vec{x}]\vec{Y} =\vec{y} \land \ldots \land [\vec{W} \gets \vec{w}]\vec{Z}=\vec{z}\}}
P_M(\vec{u}) 1_{\{ (M,\vec{u}) \sat \V =\vec{v}\}}\\
= \sum_{\{\vec{u} | (M,\vec{u}) \sat [\vec{X} \gets \vec{x}]\vec{Y} =\vec{y} \land \ldots \land [\vec{W} \gets \vec{w}]\vec{Z}=\vec{z}\}}
P_M(\vec{u}).$
\eprf

\subsection{Details on the Markov condition} 

Letting $\vec{EPa_X}$ and $\vec{XPa_X}$ respectively denote $X$'s endogenous and exogenous parents, and letting $\V = \{A,B,\ldots,K\}$, we get that $P_M(\vec{v}) = \sum_{\vec{u} \in \R(\U)} P_M(\vec{u}, \vec{v})\\ 
= \sum_{\{(\vec{xpa_A},\vec{xpa_B},\ldots,\vec{xpa_K})  \in \R(\vec{XPa_A} \times \vec{XPa_B} \times \ldots \times \vec{XPa_K})\}} \prod_{X \in \V} P_X(x | \vec{epa_X},\vec{xpa_X}) P_{\U}(\vec{xpa_X})\\
= \prod_{X \in \V} ( \sum_{\{\vec{xpa_X} \in \R(\vec{XPa_X})\}} P_X(x | \vec{epa_X},\vec{xpa_X}) P_{\U}(\vec{xpa_X}))$, where each $(x,\vec{epa_X}) \subseteq \vec{v}$, and $(\vec{xpa_A},\vec{xpa_B},\ldots,\vec{xpa_K})=\vec{u}$. (The latter means we are assuming that each $U \in \U$ is a parent of some $X \in \V$, but this is obviously without loss of generality.)

\subsection{Details on the well-definedness of Definition \ref{def:pc}}

\lem\label{lem:well} Given a PNSEM $M$ and $\vec{Y},\vec{X}, \ldots, \vec{Z},\vec{W} \subseteq \V$, then for all values $\vec{y},\vec{x},\ldots,\vec{z},\vec{w} \in \R(\vec{Y}),\R(\vec{X}),\ldots,\R(\vec{Z}),\R(\vec{W})$, $P_M(\vec{y}_{\vec{x}}, \ldots, \vec{z}_{\vec{w}} | \vec{u},\vec{v})$ is a well-defined probability distribution. 
\elem

\prf
First we show that each factor of the form $P_M(\vec{y}_{\vec{x}} | \vec{u},\vec{v})$ is a well-defined probability distribution. Let $\vec{Z}=\V - \vec{Y}$. Then $P_M(\vec{y}_{\vec{x}} | \vec{u},\vec{v}) = \sum_{\{\vec{z} \in \R(\vec{Z})\}} P_{(M^{(\vec{u}, \vec{v})})_{\vec{X} \gets \vec{x}}}(\vec{Y}=\vec{y},\vec{Z}=\vec{z} | \U=\vec{u})$. Therefore it suffices to show that $P_{(M^{(\vec{u}, \vec{v})})_{\vec{X} \gets \vec{x}}}(\V=\vec{v}^* | \U=\vec{u})$ is a well-defined probability distribution. This follows directly from the fact that it is the joint distribution of a PNSEM, and those are well-defined per assumption of satisfying the Markov condition. Concretely, we have that: 

$P_{(M^{(\vec{u}, \vec{v})})_{\vec{X} \gets \vec{x}}}(\V=\vec{v}^* | \U=\vec{u})= \prod_{\{Y \in \V \setminus \vec{X}\}} P^{(\vec{epa_Y},\vec{xpa_Y},y)}_Y(y^* | \vec{epa_Y}^*,\vec{xpa_Y}) \prod_{\{X \in \vec{X}\}}P^{alt}_X(x^*)$. 

Here $P^{alt}_X(X)$ is the probability distribution that returns $1$ if $X=x$ for the unique value $x$ such that $x \in \vec{v}$ and $0$ otherwise. Furthermore, note that for each $Y  \in \V \setminus \vec{X}$, $P^{(\vec{epa_Y},\vec{xpa_Y},y)}_Y(Y | \vec{Epa_Y},\vec{Xpa_Y})$ defines a family of conditional probability distributions over $\R(Y)$, as required.

Second, note that the product of the probability distributions of independent variables is also a probability distribution. So by {\em defining} $P_M(\vec{y}_{\vec{x}}, \ldots, \vec{z}_{\vec{w}} | \vec{u},\vec{v})$ as $P_M(\vec{y}_{\vec{x}} | \vec{u},\vec{v}) \ldots P_M(\vec{z}_{\vec{w}} | \vec{u}, \vec{v})$, we are {\em choosing} to define the counterfactual variables $\vec{y}_{\vec{x}}, \ldots, \vec{z}_{\vec{w}}$ as being mutually independent conditional on a world $(\vec{u},\vec{v})$. (As the joint distribution of these variables had no anterior meaning in the context of PNSCMs, from a purely mathematical point of view we are free to choose to define it in whatever way we want. That the definition is also a good one is of course what I try to argue in the paper.)

Third, we show that $P_M(\vec{y}_{\vec{x}}, \ldots, \vec{z}_{\vec{w}}, \vec{u},\vec{v}) = P_M(\vec{y}_{\vec{x}}, \ldots, \vec{z}_{\vec{w}} | \vec{u},\vec{v}) P_M(\vec{u}, \vec{v})$, and thus the definition of $P_M(\vec{y}_{\vec{x}}, \ldots, \vec{z}_{\vec{w}})$ is simply an application of the law of total probability.

$P_M(\vec{y}_{\vec{x}}, \ldots, \vec{z}_{\vec{w}}, \vec{u},\vec{v})\\
 = \sum_{\{\vec{u}',\vec{v}'\}} P_M(\vec{y}_{\vec{x}} | \vec{u}',\vec{v}') \ldots P_M(\vec{z}_{\vec{w}} | \vec{u}', \vec{v}') P_M(\vec{u}, \vec{v} | \vec{u}', \vec{v}') P_M(\vec{u}', \vec{v}') \\
 = P_M(\vec{y}_{\vec{x}} | \vec{u},\vec{v}) \ldots P_M(\vec{z}_{\vec{w}} | \vec{u}, \vec{v}) P_M(\vec{u}, \vec{v})\\
  = P_M(\vec{y}_{\vec{x}}, \ldots, \vec{z}_{\vec{w}} | \vec{u},\vec{v}) P_M(\vec{u}, \vec{v})$
 \eprf

\subsection{Details on Example \ref{ex:comp}}

\begin{example*}{\bf \ref{ex:comp}} Imagine again our patient from before, except that we now also consider it possible that the patient recovers even when untreated. So we get: $P_Y(y | x)=p$ and $P_Y(y | x')=q$ and $P_X(x)=r$ for some $0<p,q,r<1$, where $Y=y$/$Y=y'$ represents recovery/death and $X=x$/$X=x'$ represents treatment/no treatment. Computing the PNS gives:
$P_M(y_x,y'_{x'})=\sum_{\{x'',y''\}} P_M(y_x,y'_{x'} | x'',y'') P_M(x'',y'') = \sum_{\{x'',y''\}} P_M(y_x | x'',y'') P_M(y'_{x'} | x'',y'') P_M(x'',y'')\\ =
P_M(y_x | x,y) P_M(y'_{x'} | x,y) P_M(x,y) + P_M(y_x | x,y') P_M(y'_{x'} | x,y') P_M(x,y') \\
+ P_M(y_x | x',y) P_M(y'_{x'} | x',y) P_M(x',y) + P_M(y_x | x',y') P_M(y'_{x'} | x',y') P_M(x',y')\\= 1P_M(y'_{x'}) P_M(x,y) + 0 P_M(y'_{x'}) P_M(x,y') + P_M(y_x) 0 P_M(x',y) + P_M(y_x) 1 P_M(x',y') =\\
P_M(y'_{x'}) P_M(x,y) + P_M(y_x)P_M(x',y') = (1-q)rp+p(1-r)(1-q)=p(1-q)=P_Y(y | x)P_Y(y' | x')$
\end{example*}

\end{document}